\documentclass[10pt,twocolumn,letterpaper]{article}

\usepackage{graphicx}
\usepackage{amsmath,amsthm}
\usepackage{amssymb}
\usepackage{booktabs}
\usepackage{caption}
\usepackage{algorithm}
\usepackage{algorithmic}
\usepackage{multirow}
\newtheorem{theorem}{Theorem}
\usepackage{bm,url}
\usepackage[misc]{ifsym}
\date{}

\usepackage[pagebackref,breaklinks,colorlinks]{hyperref}

\usepackage[capitalize]{cleveref}
\crefname{section}{Sec.}{Secs.}
\Crefname{section}{Section}{Sections}
\Crefname{table}{Table}{Tables}
\crefname{table}{Tab.}{Tabs.}

\newtheorem{definition}{Definition}

\newtheorem{proposition}{Proposition}

\newcommand{\A}{\mathbf{A}}
\newcommand{\B}{\mathbf{B}}

\newcommand{\X}{\mathbf{X}}
\newcommand{\x}{\mathbf{x}}
\newcommand{\y}{\mathbf{y}}
\newcommand{\W}{\mathbf{W}}
\newcommand{\Y}{\mathbf{Y}}

\begin{document}

\title{\vspace{-2cm} \textbf{Learning Robust and Lightweight Model through Separable Structured Transformations}}

\author{
\textbf{Xian Wei$^\ast$}, \textbf{Yanhui Huang$^1$}, \textbf{Yangyu Xu$^2$}, \textbf{Mingsong Chen$^3$}, \\
\textbf{Hai Lan$^4$}, \textbf{Yuanxiang Li$^5$}, \textbf{Zhongfeng Wang$^6$}, \textbf{Xuan Tang$^7$}\\
% East China Normal University\\
{\tt\small xian.wei@tum.de}
\thanks{Corresponding Author} , 
{\tt\small huangyanhuind@gmail.com$^1$}, 
{\tt\small xuyangyu20@mails.ucas.ac.cn$^2$}, 
{\tt\small mschen@sei.ecnu.edu.cn$^3$},\\
{\tt\small lanhai09@fjirsm.ac.cn$^4$}, 
{\tt\small yuanxli@sjtu.edu.cn$^5$}, 
{\tt\small zfwang@nju.edu.cn$^6$}, 
{\tt\small 2265275624@qq.com$^7$}
% \and
% Yanhui Huang\\
% Fuzhou University\\
% {\tt\small huangyanhuind@gmail.com}
% \and
% Yangyu Xu\\
% University of Chinese Academy of Sciences\\
% {\tt\small xuyangyu20@mails.ucas.ac.cn}
% \and
% Mingsong Chen\\
% East China Normal University\\
% {\tt\small mschen@sei.ecnu.edu.cn}
% \and
% Hai Lan\\
% Fujian Institute of Research on the Structure of Matter, Chinese Academy of Sciences\\
% {\tt\small lanhai09@fjirsm.ac.cn}
% \and
% Yuanxiang Li\\
% Shanghai Jiao Tong University\\
% {\tt\small yuanxli@sjtu.edu.cn}
% \and
% Zhongfeng Wang\\
% Nanjing University\\
% {\tt\small zfwang@nju.edu.cn}
% \and
% Xuan Tang\\
% East China Normal University\\
% {\tt\small 2265275624@qq.com}
}
\maketitle

%%%%%%%%% ABSTRACT
\begin{abstract}
With the proliferation of mobile devices and the Internet of Things, deep learning models are increasingly deployed on devices with limited computing resources and memory, and are exposed to the threat of adversarial noise.
Learning deep models with both \textbf{lightweight} and \textbf{robustness} is necessary for these equipments. 
However, current deep learning solutions are difficult to learn a model that possesses these two properties without degrading one or the other.
As is well known, the fully-connected layers contribute most of the parameters of convolutional neural networks. We perform a separable structural transformation of the fully-connected layer to reduce the parameters, where the large-scale weight matrix of the fully-connected layer is decoupled by the tensor product of several  separable small-sized matrices. Note that data, such as images, no longer need to be flattened before being fed to the fully-connected layer, retaining the valuable spatial geometric information of the data.
Moreover, in order to further enhance both \textbf{lightweight} and \textbf{robustness}, we propose a joint constraint of sparsity and differentiable condition number, which is imposed on these separable matrices. We evaluate the proposed approach on \textbf{MLP}, \textbf{VGG-16} and \textbf{Vision Transformer}.
The experimental results on datasets such as ImageNet, SVHN, CIFAR-100 and CIFAR10 show that we successfully reduce the amount of network parameters by 90\%, while the robust accuracy loss is less than 1.5\%, which is better than the SOTA methods based on the original fully-connected layer.
Interestingly, it can achieve an overwhelming advantage even at a high compression rate, e.g., $200$ times.
\end{abstract}

%%%%%%%%% BODY TEXT
\section{Introduction}
\label{sec:01}
%%%%%%%%%%
\begin{figure}[t]
	\centering
	\includegraphics[width=0.9\linewidth]{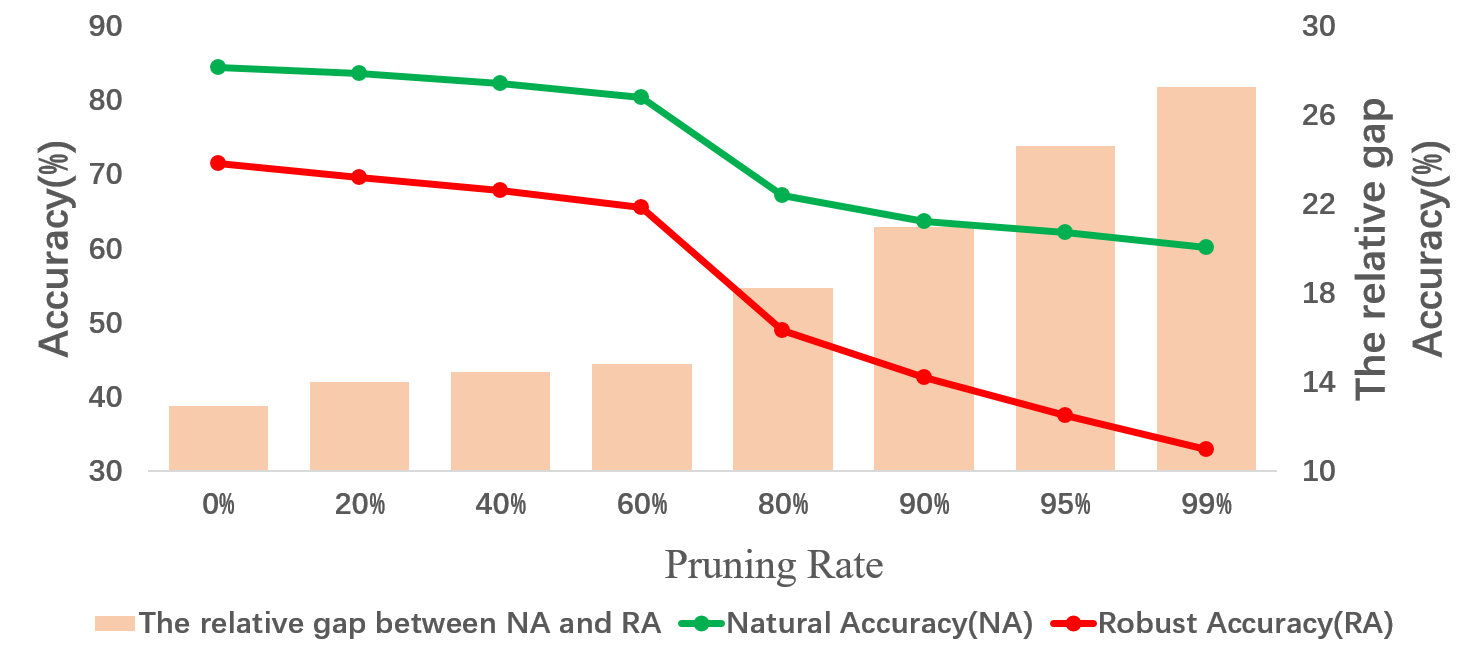}
	\caption{Natural accuracy and robust accuracy with different pruning rates. Adversarial examples are generated by the FGSM attack. This example of using ResNet-18 for classification on the CIFAR-$10$ dataset shows that pruning can reduce both \textbf{NA} and \textbf{RA}, but the drop of the latter is larger.
		\vspace*{-4mm}
	}
	\label{fig: motivation}
\end{figure}
%%%%%%%%%%%%%%%%%%%%%%
Despite Deep Neural Networks (DNNs) have made impressive achievements on machine learning tasks from computer vision to speech recognition and natural language processing \cite{chollet2017xception_cvpr, chen2018deeplab_pami, he2017wider_nips},
DNNs are vulnerable when it encounters the specially-crafted adversarial examples \cite{goodfellow2014explaining_iclr, kurakin2016adversarial_scale, xu2019structured_iclr}. 
This problem provokes people to worry about the potential risk of the DNNs and attracts the researchers' interest in the adversarial robustness of models. 
Current research of adversarial robustness mainly focuses on the standard DNNs, such as VGG \cite{simonyan2014very_vgg16}, ResNet \cite{he2016deep_cvpr_resnet} and EfficientNet \cite{tan2019efficientnet_icml}.
On the contrary, most researchers have not placed required emphasis on the robustness of lightweight networks. 
Lightweight DNNs are widely equipped in embedded devices due to the limitation of their computation, memory and power resources \cite{han2015learning_nips, khrulkov2020tensorized_iclr, yu2020search}. 
The vulnerability of lightweight DNNs
is unacceptable and dangerous for embedded systems that need high level of robustness.
For instance, the misclassification of traffic signs by self-driving cars
could lead to traffic accidents.
Therefore, for the embedded systems, it is necessary to construct lightweight DNN models that are robust with respect to various perturbations.
%%%%%%%%%%%%%%%%%%%%

However, recent research has shown that it is challenge to achieve accuracy, lightweight, %compression rate 
and robustness at high level simultaneously
\cite{wang2018adversarial_robust_iclr, guo2018sparse_nips, zhao2019_tocompress_mlsys}. 
On the one hand, most of the existing lightweight technologies, such as pruning \cite{han2015learning_nips, yang2019ecc_cvpr} and low-rank based factorization \cite{denton2014exploiting_nips, khrulkov2020tensorized_iclr}, tend to either decrease the rank of parameter matrix or cause an ill-conditioned matrix,
which may result in the models being vulnerable to perturbations from the environment \cite{horn1994topics_book}.
An example shown in Fig.~\ref{fig: motivation} demonstrates that 
 the more the network is pruned, the more the robust accuracy is lost.
% as the more A prunes, the more accuracy of B is lost
%
On the other hand, the strategies focused on robust training are likely to limit the success of achieving model lightweight, 
as deep and wide model capacities contribute to the robustness \cite{madry2017towards_iclr18, zhang2019theoretically_icml}.  
With the goal of solving the natural contradiction between lightweight and robustness, this paper designs a model that is both lightweight and robust, in which the fully-connected layer is replaced by several separable transformation matrices through the Kronecker product with joint constraints.
% \romannumeral 1) 
First of all, in order to reduce the parameters while avoiding the parameter matrix tends to low rank,
%avoid the parameter matrix tends to low rank during the compression process,
the fully-connected layer is converted into several separable small matrices through the Kronecker product.
This leads to the significant reduction in model parameters, however, the dimension of the parameter matrix of the fully-connected layer is maintained. In other words, this transformation hardly affects the expressiveness of features.
%\romannumeral 2) 
Secondly, we imposed a sparsity constraint on all underlying separable small matrices.
% This allows the sparsity constraint to be imposed for each separable transformation.
Moreover, we propose a differentiable constraint on the condition number of the parameter matrix, which structurally strengthens the robustness of the model. % , including adversarial attacks. 
% \romannumeral 3) 
Finally, we incorporate the proposed separable transformations associated with joint constraints into the framework of adversarial training, with the goal of further enhancing the model robustness against imperceptible but deliberately designed adversarial noise.

\section{Related Work}
\label{sec:02}
\paragraph{Adversarial Attack and Defense.}
% 现有的工作中，对于对抗性鲁棒性是如何获得的。最后引入条件数约束：条件数约束下的良态矩阵对于微小扰动具有一定的抵抗性【引用论文】。这与对抗性攻击的前提不谋而合，因此……
%%%%%%%%%%%%%%%%%%%%%%%%%%%%%%%%%%%%%%%%%%%%%%%%%%%%%%%%%%%%%%%%%%%%%%%%%%
% Szegedy首先发现，在大数据集上训练的模型都容易受到对抗性样本的攻击。Goodfellow探讨了对抗性样本存在和可泛化的潜在原因。此外，他们提出了两种在白盒环境下生成对抗样本的方法，即快速梯度法(FGM)和快速梯度符号法(FGSM)。在此之后，大量对抗攻击算法被提出，例如CW JSMA BIM PGD等等等。
Szegedy \textit{et al.} \cite{szegedy2013intriguing} first discovered that models trained on huge datasets are vulnerable to adversarial examples.
Goodfellow \textit{et al.} \cite{goodfellow2014explaining_iclr} explored the underlying reasons for the existence and generalisability of adversarial examples.
Additionally, they presented two methods to generate adversarial examples in a white-box setting, the fast gradient method (FGM) and the fast gradient sign method (FGSM).
After that, a large number of adversarial attack algorithms have been proposed, such as C\&W Attack \cite{carlini2017towards_CW}, DeepFool \cite{moosavi2016deepfool_cvpr}, 
the Jacobian-based Saliency Map (JSMA) \cite{papernot2016limitations_eurosp}, 
the Basic Iterative Method (BIM) \cite{kurakin2016adversarial_scale}, 
the Projected Gradient Descent attack (PGD) \cite{madry2017towards_iclr18} and
the structured adversarial attack \cite{xu2019structured_iclr}. 
%%%%%%%%%%%%%%%%%%%%%%%%%%%%%%%%%%%%%%%%%%%%%%%%%%%%%%%%%%%%%%%%%%%%%%%%%%
% 同时，大量的对抗防御方法被提出。包括对抗训练、整体训练、梯度掩蔽和模糊梯度识别、添加随机性、用插值函数代替输出激活，用NAS技术提升模型鲁棒性和防御蒸馏。
% 其中对抗训练是最有效最可靠的方法之一。我们将我们的方法纳入对抗性训练的框架中，使模型具有更强的鲁棒性
Meanwhile, plenty of adversarial defense methods have also been proposed, including
adversarial training \cite{madry2017towards_iclr18}, 
ensemble training \cite{tramer2017ensemble_iclr}, 
gradient masking \cite{papernot2017practical_accs}, 
obfuscated gradients identification \cite{athalye2018obfuscated_iclr}, 
adding randomness \cite{xie2017mitigating_iclr},
replacing the output activation by an interpolating function \cite{wang2019adversarial_activation, zhang2018efficient_nips},
using Neural Architecture Search (NAS) technology to improve model robustness \cite{guo2020meets},
and defensive distillation \cite{papernot2016distillation_sp, papernot2016limitations_eurosp}.
% Among them, adversarial training is one of the most effective and reliable strategies.
% We incorporate our method into the framework of adversarial training to make the model have stronger adversarial robustness.
 
% Our method is also built on the AT framework, while we focus on the condition number to explore 
%%%%%%%%%%%%%%%%%%%%%%%%%%%%%%%%%%%%%%%%%%%%%%%%%%%%%%%%%%%%%%%%%%%%%%%%%%%%
\paragraph{Lightweight and efficient methods.}
%%%%%%%%%%%%%%%%%%%%%%%%%%%%%%%%%%%%%%%%%%%%%%%%%%%%%%%%%%%%%%%%%%%%%%%%%%%
% 现有模型的轻量化工作主要分为两类………………在上述方法中，都未考虑到轻量化模型在对抗鲁棒性上的表现，而****在***文章中，首次提出将这两者融合起来……因此，本文在**的基础上，提出了一种**的框架，从**角度对轻量化模型的鲁棒性进行提升。
%%
% 构建轻量级深度学习模型有两种主要方法，一种是压缩过度参数化的深度模型，然后通过剪枝、分解或蒸馏预先训练的权值来压缩成更小的网络，同时保持有竞争力的准确性，另一种是直接训练小网络。Yu使用神经体系结构搜索(NAS)技术来寻找和构建小但有竞争力的模型，而Iandola提出了SqueezeNet体系结构，使用更小的滤波器、更少的输入通道和延迟的下行采样。
%
Most lightweight networks are obtained by either compressing existing over-parameterized DNN models, called model compression, or designing small networks directly.
The typical model compression techniques include
% There are two main approaches to building lightweight deep learning models by compressing an over-parameterized deep model which is then compressed into a smaller network by 
pruning \cite{han2015learning_nips, yang2019ecc_cvpr}, factorizing \cite{denton2014exploiting_nips, yu2017compressing_cvpr, khrulkov2020tensorized_iclr}, 
quantizing \cite{wu2016quantized_cvpr, gennari2020finding, yu2020search},
or distilling \cite{hinton2015distilling_nips, polino2018model_ICLR} the pre-trained weights while maintaining competitive accuracy.
% , or by training small networks directly.
%
To avoid the pre-training over-parameterized large models, one could also focus on directly building small and efficient network architectures that could be trained from scratch.
% 	\cite{iandola2017squeezenet_iclr, howard2017mobilenets, chollet2017xception_cvpr, sandler2018mobilenetv2_CVPR}.
	%
For example, SqueezeNet \cite{iandola2017squeezenet_iclr} proposed lightweight CNN architectures by using smaller-sized filters, less input channels, and delayed downsampling.
	%reduced the parameters of CNN by replacing $3\times 3$ filters with $1\times 1$ filter, decreasing the numbers of input channels to $3\times 3$ 
	%
	%Another popular approach  is to employ factorization operations such as depthwise separable convolution for lightweight convolution layers. 
Many notable deep learning models such as Xception \cite{chollet2017xception_cvpr}, MobileNet V1\cite{howard2017mobilenets},  MobileNet V2 \cite{sandler2018mobilenetv2_CVPR},
	% cf.~\cite{howard2017mobilenets, , sandler2018mobilenetv2_CVPR}.
	%
ShuffleNet \cite{zhang2018shufflenet_cvpr} and CondenseNet \cite{huang2018condensenet_cvpr} employed  depthwise separable convolution.
% Yu \textit{et al.} use neural architectures search (NAS) technology to find and build small but competitive accuracy model  \cite{yu2020search, gennari2020finding, li2020cp}, while Iandola \textit{et al.} proposed SqueezeNet architectures by using smaller sized filters, less input channels, and delayed downsampling \cite{iandola2017squeezenet_iclr}.
%%%%%%%%%%%%%%%%%%%%%%%%%%%%%%%%%%%%%%%%%%%%%%%%%%%%%%%%%%%%%%%%%%%%%%%%%%%%%
%%%%%%%%%%%%%%%%%%%%%%%%%%%%%%%%%%%%%%%%%%%%%%%%%%%%%%%%%%%%%%%%%%%%%%%%%%%
% 然而，对于上述方法，他们没有考虑轻量级模型的对抗鲁棒性的重要性。
%However, Regarding the above approaches, they don't 
All these methods pay little attention to the robustness of lightweight models. 
% Even more severe, some of the techniques cause the decrease of the model's adversarial robustness \cite{horn1994topics_book}.
%%%%%%%%%%%%%%%%%%%%%%%%%%%%%%%%%%%%%%%%%%%%%%%%%%%%%%%%%%%%%%%%%%%%%%%%%%%
%更严重的是，一些技术会降低模型的对抗鲁棒性，如剪枝和基于低秩的因子分解。这些方法往往使矩阵秩减少或病态矩阵引用，这可能导致模型容易受到来自对抗性样本的扰动。
% Even more severe, some of the techniques cause the decrease of the model's adversarial robustness, such as pruning and low-rank-based factorizing.  Those methods tend to make matrix rank decreasing or ill-conditioned matrices \cite{horn1994topics_book}, which may result in the models being vulnerable to perturbations from the adversarial sample.
%%%%%%%%%%%%%%%%%%%%%%%%%%%%%%%%%%%%%%%%%%%%%%%%%%%%%%%%%%%%%%%%%%%%%%%%%%%%%%%%%%%
\paragraph{Robust and lightweight learning method.}
%最近的一些工作试图通过将模型压缩技术合并到对抗防御框架中来构建既健壮又轻量级的模型 ，例如剪枝 量化 蒸馏等 。
Some recent works have attempted to build models that are both robust and lightweight by incorporating model compression techniques into the adversarial defense framework.
%
%Sehwag将修剪过程作为对抗损失目标中的经验风险最小化问题进行阐述
Sehwag \textit{et al.} \cite{sehwag2020pruning_adversarial} formulated the pruning process as an empirical risk minimization problem within adversarial loss objectives.
%
%Madaan提出了一种通过剪枝漏洞高的潜在特征来抑制漏洞的方法。
Madaan \textit{et al.} \cite{madaan2019adversarialNeuralPruning} proposed 
to suppress vulnerability by pruning the latent features with high vulnerability.
% 
%Ye 将权值修剪引入到与最小最大优化相关联的对抗训练框架中，在保持鲁棒性的同时实现模型压缩
Ye \textit{et al.} \cite{Ye2019Adversarial_both_ICCV} incorporated the weight pruning into the framework of adversarial training associated with min-max optimization, to enable model compression while preserving robustness.
% 
%除了权值修剪，Lin 提出了一种新的防御量化方法，通过在量化过程中控制神经网络的Lipschitz常数。作者在XX中开发了蒸馏方法，通过最小化教师在自然图像上的输出与学生在对抗性图像上的输出之间的差异，生成健壮的学生网络。
Other than weight pruning, Lin \textit{et al.} \cite{lin2019defensive_iclr} proposed a novel defensive quantization method by controlling the neural network’s Lipschitz constant during quantization.
The authors developed in \cite{goldblum2019adversarially_distillation} distillation methods that produce robust student networks
by minimizing discrepancies between the outputs of a teacher on natural images and the outputs of a student on adversarial images.
%
%最近，Gui 提出了一个以剪枝、分解和量化为约束的对抗训练统一框架
Recently, Gui \textit{et al.} \cite{gui2019model_nips} proposed a unified framework for adversarial training with pruning, factorization and quantization being the constraints.
%上述方法将模型压缩与对抗防御策略相结合，主要关注于实现对对手的健壮模型
The aforementioned methods combined the model compression and adversarial defense strategies, with the main focus on achieving adversarially robust models, without addressing inherent contradictions between model compression and model robustness.  
\section{Learning Robust and Lightweight Transformation with Separable Structures}
\label{sec:03}
Most deep learning models consist of dozens levels of linear transformations and activation functions, as follows, 
%associated with a task-driven loss function. 
% , pooling layers
%Considering that most pooling and activation functions are predefined with fixed policies, 
%the aforementioned learning capabilities are highly dependent on the following linear transformation, 
%
\begin{equation}
	\label{eq_basic_linear_system}
	\y = \W \x +\bm{b}, \;\;\;  \mathbf{h}  = \sigma(\y)
	% \y = \W\circ \x %\y = \W\x +\bm{b}
\end{equation}
%
%%
%\begin{align}
%	\label{eq_basic_linear_system}
%	\y & = \W \x, \\
%	\label{eq_basic_linear_system_activation}
%	\mathbf{h} & = \sigma(\y)
%	% \y = \W\circ \x %\y = \W\x +\bm{b}
%\end{align}
%%
%with $\x$, $\y$, $\bm{b}$, $\W$ being an input signal, the output response, the bias and the corresponding linear transformation.
where $\x$, $\bm{b}$, $\W$ being an input signal, the bias term, and the corresponding linear transformation matrix, respectively. % and a product operator. 
$\mathbf{h}$ is the output response, $\sigma(\cdot)$ denotes an activation function.
% $\circ$
If $\x$, $\mathbf{h}$ lie in 1-dimensional (1D) tensor space,
%The $1-$dimensional (1D) model in 
%the system 
Eq.~\eqref{eq_basic_linear_system} covers prominent deep learning models
like CNNs, RNNs and auto-encoders. % \cite{Goodfellow_Bengio2016_Book}. 
For example, each row of $\W$ could be a reshaped convolutional filter of CNNs with
$\W$ being a collection of matrices that contains all convolutional filters in certain layer \cite{zhangAccelerating_pami16}.
For the fully-connected (FC) layer of CNNs or auto-encoders, $\W$ is a common linear transformation matrix.
% and $d$ could be the number of filters in the layer of CNN \cite{zhangAccelerating_pami16},
%Each row of $\W$ denotes the reshaped form of a $k\times k \times c$ filter with the bias appended.
%
Generally, given a multi-dimensional (MD) input signal $\mathcal{X}$,
%which could be an image or a patch with one channel,
one common way in a DNN is to
reshape $\mathcal{X}$ into a 1-dimensional tensor  $\mathbf{x}$  and feed it to the system Eq.\eqref{eq_basic_linear_system}.
%given a one dimensional (1D) signal $\x \in \mathbb{R}^m$,
This may result in the loss of spatial geometry structure information and expensive computation costs when encountering high-dimensional input signal.
%

% In order to develop a learning model that is both \textit{robust} and \textit{lightweight},  
Learning a both robust and lightweight model,
we want the system in Eq.~\eqref{eq_basic_linear_system} to have the following properties:
\romannumeral 1) $\W$ has a small amount of parameters, and its entries are sparse in a reduced data format. % \cite{choudhary2020comprehensive_air}.%have smaller storage burden.
%		or quantized
% \romannumeral 2) $\W$ is a well-conditioned matrix with robustness to permutations. % \cite{cline1979estimate_siam}.
\romannumeral 2) $\mathbf{h}$ is robust to small changes around the input data, which could be 
   achieved by learning a well-conditioned transformation $\W$ and preventing major changes from activation function.
   % the geometry of training data.
%   % preserving the data geometrical  structures. 

To address the difficulty of a learning model in attaining both properties without degrading each other,
we exploit well-conditioned matrices with small sizes that work for the system of Eq.~\eqref{eq_basic_linear_system}.
In the following, we first introduce separable structured linear transformations that have very small sizes, 
%that admits a separable structure,
%where separable means that the transformation is given by the Kronecker product of several very smaller matrices.
such that we can effectively impose the robustness and the sparsity constraints on the separable small-sized matrices. 
%To achieve this goal, in the following, we construct the model from two aspects:

%%%%%%%%%%%%%%%%%%%%%%%%%%%%%%%%%%%%%%%%%%%%%%%%%%%%%%%%%%%%%%%%%%%%%%%%%%%%
\subsection{Learning Separable Structured Transformations}
% \subsection{Learning $D$ dimensional separable matrices}
%
\label{sec:031}
We consider the general case here by assuming that the input signal is a multi-dimensional tensor.
It is worth noting that the dimension of transformation is essentially limited by memory and computing resources.
Consequently, the transformation is allowed to have a separable structure. In other words, the input tensor is replaced by the tensor product (a.k.a. Kronecker product) of several small-sized matrices.

In order to establish the multiplication of a higher-order signal tensor by \textit{separable} matrices, we define the
$n-$mode product as follows, by referring to the concept of tensor operation in \cite{van2000ubiquitous_jcam}.
\begin{definition}[$n-$mode product]
	The $n-$mode product of a tensor $\mathcal{X}\in \mathbb{R}^{I_1\times I_2\times \cdots \times I_N}$ 
	by a matrix $\A\in \mathbb{R}^{K_n\times I_n}$ with $n\leq N$, is denoted by 
     $\mathcal{X} \times_{n}\mathbf{A},$
	which is an $(I_1\times I_2\times \cdots \times I_{n-1}\times K_{n} \times I_{n+1}\times \cdots \times I_{N})-$tensor.
	For all $k_n = 1, \cdots, K_n$, the entries of the tensor $\mathcal{X} \times_{n}\mathbf{A}$ are defined as
	$$(\mathcal{X} \times_{n}\mathbf{A})_{i_1 i_2\cdots i_{n-1} k_n i_{n+1} i_{N}} = \sum_{i_n = 1}^{I_n}{X_{i_1 i_2\cdots i_{N}}A_{k_n i_n}}$$
	with $X_{i_1 i_2\cdots i_{N}}$, $A_{k_n i_n}$ being entries in $\mathcal{X}$ and $\A$, respectively.
\end{definition}
Let us now refer to the linear transformation in Eq.~\eqref{eq_basic_linear_system}.
 Suppose a tensor signal 
$\mathcal{X}\in \mathbb{R}^{I_1\times I_2\times \cdots \times I_N}$ is multiplied by $T$ separable transformation matrices, 
$\mathcal{A} := \{\mathbf{A}^{(1)} \in \mathbb{R}^{K_1 \times I_1}, 
\mathbf{A}^{(2)} \in \mathbb{R}^{K_2 \times I_2}, \cdots, \mathbf{A}^{(T)} \in \mathbb{R}^{K_T \times I_T}\} $. 
%we reformulate the Eq.~\eqref{eq_basic_linear_trans} and Eq.~\eqref{eq_separable_layer} as
%we formulate $\mathcal{X}$ as the $n-$mode product of 
The response $\mathcal{Y}\in \mathbb{R}^{K_1\times K_2\times \cdots \times K_N}$ is 
formulated as the $n-$mode product of $\mathcal{X}$ and these $T$ separable transformation matrices as
\begin{equation}
	\label{eq_basic_linear_trans_MD}
	%\y = \sigma \left(\W\x + \mathbf{c} \right)
	%\mathbf{Y} = \sigma\left( \mathbf{A}\mathbf{X}\mathbf{B}^{\top} + \mathbf{A} \mathbf{C} \mathbf{B}^{\top} \right) 
	%\mathcal{Y} = \sigma\left( \mathcal{X} \times_{1}\mathbf{A}^{(1)}  \times_{2}\mathbf{A}^{(2)} \times_{3} \cdots  \times_{T} \mathbf{A}^{(T)} +  \mathcal{C} \right). 
	\mathcal{Y} =  \mathcal{X} \times_{1}\mathbf{A}^{(1)}  \times_{2}\mathbf{A}^{(2)} \times_{3} \cdots  \times_{T} \mathbf{A}^{(T)}. 
\end{equation}
The increasing of $T$ reduces the computational load of the model, but it may degrade the expressiveness of the parameters \cite{qi2017multi_tpami, seibert2016learning_tsp}.
%\subsubsection{Connection with 1D model}
%Similar to the form in Eq.~\eqref{eq_separable_multi_layer}, 
Instead, the transformation in Eq.~\eqref{eq_basic_linear_trans_MD} can
be conveniently converted to the 1D model of \eqref{eq_basic_linear_system} with a structured transformation as follows:
% $\mathbf{A}^{(1)}, \mathbf{A}^{(2)}, \cdots, \mathbf{A}^{(T)}$, as
%
\begin{equation}
	\begin{aligned}
		\label{eq_separable_multi_layer_tensor}
		%\text{vec}(\mathcal{Y}) = \sigma\left( \left( \mathbf{A}^{(T)} \otimes \cdots \otimes \mathbf{A}^{(2)} \otimes \mathbf{A}^{(1)}\right) \text{vec}(\mathcal{X}) + \mathbf{c} \right) 
		\text{vec}(\mathcal{Y}) =  \left( \mathbf{A}^{(1)} \otimes \mathbf{A}^{(2)} \otimes \cdots  \otimes \mathbf{A}^{(T)}\right) \text{vec}(\mathcal{X}),
		%& = \sigma\left(  \text{vec} \left( {\A}^{(1)} {\text{vec}}^{-1}(\x) {\left(\mathbf{A}^{(T)} \otimes \cdots \otimes \mathbf{A}^{(2)}\right)}^\top  \right) + \mathbf{c}  \right),
		%& = \sigma\left(  \text{vec} \left( {\A}^{(1)} \X {\left(\mathbf{A}^{(T)} \otimes \cdots \otimes \mathbf{A}^{(2)}\right)}^\top  \right) + \mathbf{c}  \right)
	\end{aligned}
	% y = \sigma(B $\otimes$ A)x + b
\end{equation}
%
%with $\mathbf{c} = \text{vec}(\mathcal{C})$.
%with $\text{vec}(\mathcal{Y})$ and $\text{vec}(\mathcal{X})$ being the 
where the vector space isomorphism $\text{vec}: \mathbb{R}^{a\times b} \rightarrow \mathbb{R}^{a b}$
is defined as the operation that stacks the columns on top of each other,
%and $\text{vec}^{-1}: \mathbb{R}^{ab} \rightarrow \mathbb{R}^{a\times b}$, 
e.g., $\x = \text{vec}(\mathcal{X})$ and $\y = \text{vec}(\mathcal{Y})$.
%$\X = \text{vec}^{-1}(\x)$ in Eq.~\eqref{eq_separable_layer}.
%
Therein, $\otimes$ denotes the Kronecker product operator. % \cite{van2000ubiquitous_jcam}.
%, with which the Kronecker product of $\mathbf{A}$ and $\mathbf{B}$ is defined by
%\begin{equation}
%	\label{eq_Kronecker}
%	\textbf{A} \otimes \textbf{B}=
%	\begin{bmatrix}
%		A_{11}\textbf{B} & \cdots & A_{1k_1}\textbf{B}\\
%		\vdots & \ddots & \vdots\\
%		A_{a1}\textbf{B} & \cdots & A_{ak_1}\textbf{B}\\
%	\end{bmatrix}
%\end{equation}
%with $A_{ij}$ being an entry in $\A$. 
%%
We refer to a large, linear transformation matrices $\W$
that can be represented as a concatenation of smaller matrices
$\mathcal{A} := \{\mathbf{A}^{(1)}, \mathbf{A}^{(2)}, \cdots, \mathbf{A}^{(T)}\}$
as a separable transformation,
\begin{equation}
	\label{eq_matrix_kronecker_decomposition}
	\W = \mathbf{A}^{(1)} \otimes \mathbf{A}^{(2)} \otimes \cdots \otimes \mathbf{A}^{(T-1)} \otimes \mathbf{A}^{(T)},
\end{equation}
%
%with the properties \cite{graham2018kronecker_book} of 
with the property \cite{graham2018kronecker_book} of 
\begin{align}
	\begin{split}
		\label{eq_matrix_kronecker_decomposition_exchange}
		\mathbf{A}^{(1)} \otimes \mathbf{A}^{(2)} \otimes \mathbf{A}^{(3)} 
		& = (\mathbf{A}^{(1)} \otimes \mathbf{A}^{(2)}) \otimes \mathbf{A}^{(3)} \\
		& = \mathbf{A}^{(1)} \otimes ( \mathbf{A}^{(2)} \otimes \mathbf{A}^{(3)} ),
	\end{split}
\end{align}
\begin{equation}
	\label{eq_matrix_kronecker_norm}
	\|\mathbf{A}^{(1)} \otimes \mathbf{A}^{(2)} \otimes \mathbf{A}^{(3)}\|
	= \|\mathbf{A}^{(1)}\| \|\mathbf{A}^{(2)}\| \|\mathbf{A}^{(3)}\|,
\end{equation}
%
%and
%
\begin{equation}
	\label{eq_matrix_kronecker_decomposition_rank}
	\text{rank}(\W) = \text{rank}(\mathbf{A}^{(1)})\text{rank}(\mathbf{A}^{(2)})\cdots\text{rank}(\mathbf{A}^{(T)}).
\end{equation}
As shown in Eq.~\eqref{eq_basic_linear_system} and Eq.~\eqref{eq_matrix_kronecker_decomposition}, 
the overall dimension of $\W$ is $K_1 \times I_1 \times K_2 \times I_2 \times \cdots \times K_T \times I_T$.  In comparison,  $\mathcal{A}$ has a  much smaller size 
$K_1 \times I_1 + K_2 \times I_2 + \cdots + K_T \times I_T$.
In other words, the size of problem \eqref{eq_basic_linear_trans_MD}
is much more \textit{lightweight} than the size of problem \eqref{eq_basic_linear_system} given the same input signal.
\subsection{Learning Robust and Lightweight Model via Separable Transformations}
\label{sec:032}
In this subsection, we show how to incorporate  the aforementioned separable transformations of tensors 
%introduced in Section~\ref{sec:031} 
into a task-specific learning model, e.g., classification, with the proposed regularizations. 

Let $\mathcal{X}\in \mathbb{R}^{I_1\times I_2\times \cdots \times I_N}$, ${z}$ be an input tensor signal 
and the corresponding label, respectively.  
$\mathcal{X}$ together with ${z}$ follow an underlying data distribution $\mathfrak{D}$ in data space.
A well-trained classifier model is a mapping $f \colon \mathcal{X} \mapsto {z}$
%$$f \colon \mathcal{X} \mapsto y$$
with a collection of separable parameters $\mathcal{A}$, 
denoted by ${z} = f(\mathcal{A}, \mathcal{X})$.
%In the $i^{th}-$layer of the model, 
%$$ \mathcal{A}_i = \{\{\mathbf{A}^{(1)}_i, \mathbf{A}^{(2)}_i, \cdots, \mathbf{A}^{(T)}_i\}\} $$ 
%with $\mathbf{A}^{(t)}_i$ being a two-dimensional matrix defined in Section~\ref{sec:031}.
Therein, the layer of linear transformation is constructed by Eq.~\eqref{eq_basic_linear_trans_MD}.
Generally, there exists a suitable loss function $L(\mathcal{A}, \mathcal{X}, {z})$ to measure the risk of mapping $f$. 
For example, $L$ is the cross-entropy loss for a neural network $f$ \cite{madry2017towards_iclr18}.
%  $L$ is the loss function used to train the adversarial model, e.g., cross-entropy for a neural network \cite{madry2017towards_iclr18}, 
%
The goal of the model $f$ is to find parameters $\mathcal{A}$ that minimize the risk 
$\mathbb{E}_{(\mathcal{X}, {z})\sim \mathfrak{D}}[L(\mathcal{A}, \mathcal{X}, {z})]$.
% 	%
% 	\begin{equation} 
% 		\label{eq_cost_linear_transformation}
% 	\mathbb{E}_{(\mathcal{X}, {z})\sim \mathfrak{D}}[L(\mathcal{A}, \mathcal{X}, {z})].
% 	\end{equation}
% 	%
% %
Furthermore, it is expected that this model is robust to perturbations and contains as few parameters as possible.
%
%In the work, the goal is to train a model to make the mapping $f \colon \mathcal{X} \mapsto y$ 
To achieve this goal, the following develops several constraints that are imposed on parameters $\mathcal{A}$ 
defined in Eq.~\eqref{eq_basic_linear_trans_MD}.

Considering the MD tensor operation in the model, %of \eqref{eq_basic_linear_trans_MD}, 
it is difficult to directly develop constraints for the system \eqref{eq_basic_linear_trans_MD}.
However,  the mutual transformation (see Eq.~\eqref{eq_separable_multi_layer_tensor} and
Eq.~\eqref{eq_matrix_kronecker_decomposition} 
%and their properties in Eq.~\eqref{eq_matrix_kronecker_decomposition_exchange},
%Eq.~\eqref{eq_matrix_kronecker_norm} and Eq.~\eqref{eq_matrix_kronecker_decomposition_rank}
) 
between MD model and 1D model enables solving 
MD transformation problem to take advantage of 1D classic and high efficient algorithms.
In order to develop learning paradigms for further promoting \textit{lightweight} and  \textit{robustness} by using separable parameters,
we now derive the following propositions between the MD model and the 1D model,
and hence construct appropriate regularizations.

\subsubsection{Sparsity Constraint for Pruning}
\label{sec:0321}
\begin{figure}[t]
	\centering
	\includegraphics[width=0.8\linewidth]{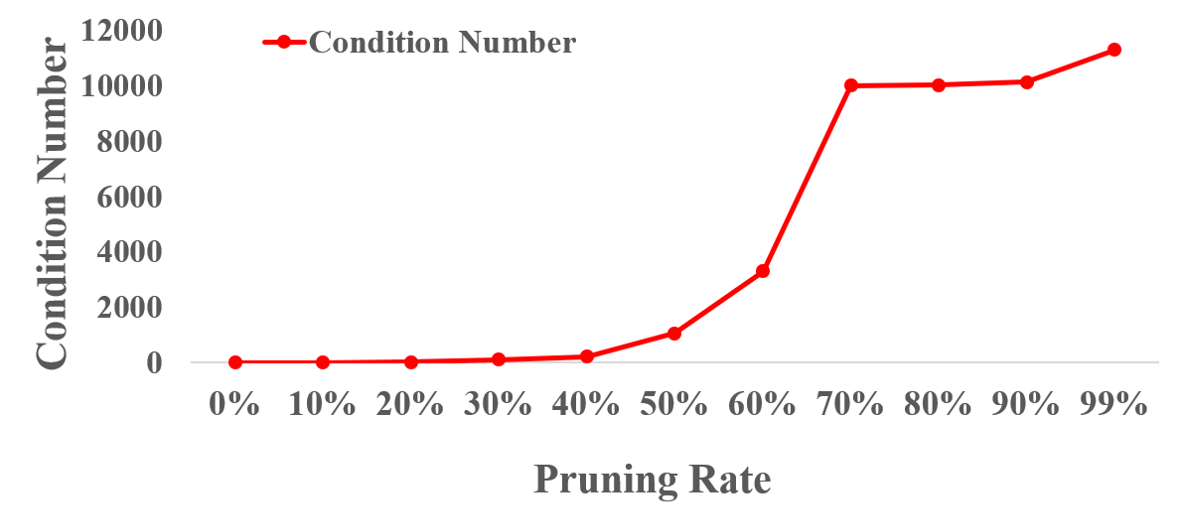}
	\caption{ The condition of parameter matrix becomes worse following the increase of pruning.
		\vspace*{-4mm}
	}
	\label{fig_condition_number_vs_sparsity}
\end{figure}

%www
% As introduced in Section~\ref{sec:03}, 
% The linear transformation with a collection of separable parameters $\mathcal{A}$
% exactly improves the \textit{lightweight}  of model.
% % will reduce the computational load of the model.
In order to further reduce the computational load and the complexity of separable parameters, in the following, 
one way is to prune the unimportant connections with small weights in training models.
\begin{definition}[$s-$sparse tensor]
	If a tensor $\mathcal{X}\in \mathbb{R}^{I_1\times I_2\times \cdots \times I_N}$ satisfies sparsity condition 
	$		\|\mathcal{X}\|_0 \leq s, \; \mathrm{for}\;\; s \in \mathbb{Z}^{+},$
% 	%
% 	\begin{equation} 
% 		\label{eq_201_sparsity_L0}
% 		\|\mathcal{X}\|_0 \leq s, \; \mathrm{for}\;\; s \in \mathbb{Z}^{+},
% 	\end{equation}
% 	%
	%According to $\ell_p$ norm \cite{ex207}, 
	we call the tensor $\mathcal{X}$ is $s$-sparse. 
	Here, $\|\mathcal{X}\|_0$ denotes the number of nonzero terms in $\mathcal{X}$.
\end{definition}
\begin{proposition}
	\label{prop:01}
	Given $\W = \mathbf{A} \otimes \mathbf{B}$ with $ \mathbf{W}\in \mathbb{R}^{K_1K_2\times I_1I_2}$,
	$ \mathbf{A}\in \mathbb{R}^{K_1\times I_1}$ being $s_{\mathbf{A}}-$sparse
	and $\mathbf{B}\in \mathbb{R}^{K_2\times I_2}$ being $s_{\mathbf{B}}-$sparse, 
	the sparsity of $\W$ is $s_{\mathbf{A}}K_1K_2 + s_{\mathbf{B}}I_1I_2 - s_{\mathbf{A}}s_{\mathbf{B}}$.
\end{proposition}
%
%As shown in (3), the n-mode product which is used in the cost function can be rewritten as a matrix-vector-product. 
%The Kronecker product of the small operators that appears in (3) can be interpreted as a large cosparse analysis operator 
%with additional structure which is applied to the vectorized signal.
%In Eq.~\eqref{eq_separable_multi_layer} and Eq.~\eqref{eq_separable_multi_layer_tensor}, 
%it has shown that the equalization between the matrix-vector-product and the $n-$mode product,
%while the latter significantly reduces the computational load and complexity.
%%%
%A large transformation matrix $\W$ can be constructed by the Kronecker product of $T$ small separable matrices,
%as shown in Eq.~\eqref{eq_matrix_kronecker_decomposition}.
%%
The model pruning is to make $T$ separable matrices $\mathcal{A}$ have few nonzero elements, while ensuring 
% the matrices set 
$\mathcal{A}$ without reducing expressiveness. 
As shown in Proposition \ref{prop:01}, the sparsity of whole 1D system is determined by the sparsity of each separable matrix. 
Therefore, a natural way for pruning is to promote the sparsity of each element in $\mathcal{A}$, given by,
\begin{equation} 
	\label{eq_sparsity}
	g(\mathcal{A}) = \sum_{t =1}^T \sum_{i_n, k_n}g\left( A^{(t)}_{i_nk_n} \right) 
\end{equation}
with $A^{(t)}_{i_nk_n}$ being an element of $\A^{(t)}$. 
%Therein, the sparsity measurement function $g$ is chosen to be separable, i.e., its evaluation is computed 
%as the sum of functions of the individual components of its argument.
For example, by appealing to the $\ell_p$ norm with $0 < p \leq 1$, % (shown in Appendix~\ref{sec:appendix_opti}), 
$g(\A^{(t)})=\|\A^{(t)}\|_p$ is used. %in Section~\ref{sec:04}.
%
% With the parse weights being obtained, the pruning is a natural way in each iteration.
Promoting the sparse structure of $\mathcal{A}$ helps its pruning in the training process,
but it may  lead to ill-conditioned matrices for $\mathcal{A}$, see Fig.~\ref{fig_condition_number_vs_sparsity}.
% tend to either decrease matrix rank or lead to ill-conditioned matrices,

%
%%%%%%%%%%%%%%%%%%%%%%%%%%%%%%%%%%%%%%%%%%%%%%%%%%%%%%%%%%%%%%%%%%%%%%%%%%
%%%%%%%%%%%%%%%%%%%%%%%%%%%%%%%%%%%%%%%%%%%%%%%%%%%%%%%%%%%%%%%%%%%%%%%%%%%%

%%%%%%%%%%%%%%%%%%%%%%%%%%%%%%%%%%%%%%%%%%%%%
\subsubsection{Moderate Condition Number}
\label{sec:0322}
The condition number of $\W$ in Eq.~\eqref{eq_basic_linear_system} determines how much a small perturbation of the input $\x$ can change the output $\y$.
% % It  sensitive is $f$ to perturbations to the input $\x$
% It is commonly used to measure how sensitive a function is to changes or errors in the input, 
% and how much error in the output results from an error in the input. 
% %The latter causes the vanishing and exploding gradient problem.
A matrix with the condition number being close to one is said to be "\textit{well-conditioned}", 
while a matrix with a large condition number is said to be ``\textit{ill-conditioned}'', which 
causes the vanishing and exploding gradient problem 
\cite{demmel1997applied_siam_book}. 
% sinha2018neural_EKDD, graham2018kronecker_book, jose2018kronecker_icml
% cline1979estimate_siam
Both reached the limit, a unitary matrix has a condition number of $1$, 
while a low-rank matrix has a condition number equal to infinite.
The unitary transformation may degrade the expressiveness of learning models, but the low-rank transformation is sensitive to perturbations.
Therefore, the transformation with a moderate condition cumber is necessary for many linear learning models.

%
%The ratio C of the largest to smallest singular value in the singular value decomposition of a matrix. 
% Condition numbers express the worst-case sensitivity of the solution of a problem to small perturbations in data A and b ($Ax = b$). 
% The product of condition number times backward error provides an approximate upper bound on the error in a computed solution.
% one should think of the condition number as being (very roughly) the rate at which the solution x will change with respect to a change in b. 
%
%
\begin{definition}[$\ell_2-$norm condition number \cite{demmel1997applied_siam_book}]
\label{eq_condition_number_define}
	The $\ell_2-$norm condition number of a full-rank matrix $\A \in \mathbb{R}^{K\times I}$ is defined as
	$	\kappa(\A) = \dfrac{\sigma_{\max} (\A)}{\sigma_{\min} (\A)},$
% 	%
% 	\begin{equation} 
% 	\label{eq_condition_number_define}
% 	\kappa(\A) = \dfrac{\sigma_{\max} (\A)}{\sigma_{\min} (\A)},
% 	%\kappa_2(\A) = \|\A\|_2 \cdot \|\A^{+}\|_2,
% 	\end{equation}
% 	%
	where $\sigma_{\max} (\A)$ and $\sigma_{\min} (\A)$ are maximal and minimal 
	singular values of $\A$. %, respectively.
	% 	%
	% 	where $\A^{+}$ is the pseudo inverse of the matrix $\A$ and the norm is defined by
	% 	$$\|\A\|_2 = {\max}_{\x \neq 0} \frac{\|\A\x\|}{\|\x\|} = \sigma_1 (\A), \;\; \|\A^{+}\|_2 = 1/\sigma_{p} (\A)$$
	% 	with $p = \textit{min}\{a, b\}$, $\sigma_1 (\A)$ and $\sigma_p (\A)$ being the largest and the smallest singular values of $\A$, respectively.
\end{definition}
A robust linear system like Eq.~\eqref{eq_basic_linear_system} often
requires the transformation matrix is as ``\textit{well-conditioned}'' as possible.
Considering the equal conversion between a linear tensor system \eqref{eq_basic_linear_trans_MD} 
and its 1D version in Eq.~\eqref{eq_separable_multi_layer_tensor},
the perturbation analysis for the problem \eqref{eq_basic_linear_trans_MD} can be achieved by 
observing the Kronecker product linear system \eqref{eq_separable_multi_layer_tensor} 
\cite{xiang2005perturbation_jcam}.
%ling2015perturbation_jcam
Therefore, let all separable matrices are full rank, with the properties in 
Eq.~\eqref{eq_matrix_kronecker_decomposition_exchange} and Eq.~\eqref{eq_matrix_kronecker_norm}, 
the condition number for the Kronecker product linear system \eqref{eq_separable_multi_layer_tensor} 
associated with Kronecker decomposition \eqref{eq_matrix_kronecker_decomposition} is deduced as
\begin{equation} 
\label{eq_condition_number}
\kappa(\W) = \kappa(\mathbf{A}^{(1)} ) \kappa(\mathbf{A}^{(2)} ) \cdots \kappa(\mathbf{A}^{(T)} ).
\end{equation}
The proof is given in Supplementary Material.
%Appendix. 

According to Eq.~\eqref{eq_matrix_kronecker_norm}, Eq.~\eqref{eq_matrix_kronecker_decomposition_rank}
and Eq.~\eqref{eq_condition_number}, properties of the whole tensor system \eqref{eq_basic_linear_trans_MD} 
are heavily dependent on the construction of each separable matrix $\{\mathbf{A}^{(t)}\}$.
Hence, in order to improve the robustness of defined tensor system, it is necessary to  
approximately hold a moderate condition number for each separable matrix $\{\mathbf{A}^{(t)}\}$.
%, which results in  developing to minimize the gap between the maximal and the minimal sigular values.
%
According to the definition of $\ell_2-$norm condition number,  
one viable way is to develop some smooth measures to prevent singular values from being essentially small and extremely large.

Let $\{\sigma_i\}_{i = 1}^k, k = \min\{a, b\}$ denote the singular values of a separable matrix $\mathbf{A}^{(t)} \in \mathbb{R}^{a\times b}$ 
in decreasing order of magnitude, and $\sigma_{\max}(\mathbf{A}^{(t)})$ denotes the largest one.
It is known that $\|\mathbf{A}^{(t)}\|_F^2 = {\sum_{i = 1}^{k}}\sigma_i^2 \geq \sigma_{\max}(\mathbf{A}^{(t)})^2$. 
Thus, imposing a penalty as in Eq.~\eqref{eq_main_func_pemalty_Frob} could result in a small $\sigma_{\max}(\mathbf{A}^{(t)})$.
\begin{equation}
\label{eq_main_func_pemalty_Frob}
\resizebox{.5\hsize}{!}{$
\rho({\mathcal{A}}) = \dfrac{1}{2 Tk^2} \sum_{t = 1}^{T}\|\mathbf{A}^{(t)}\|_F^2 $},
\end{equation}
On the other hand, $\mathbf{A}^{(t)}$ is expected to be full rank, and the Gram matrix ${\mathbf{A}^{(t)}}^\top \mathbf{A}^{(t)}$ is positive definite,
which implies $\text{det}({\mathbf{A}^{(t)}}^\top \mathbf{A}^{(t)}) > 0$. 
Recalling that $\text{det}({\mathbf{A}^{(t)}}^\top \mathbf{A}^{(t)}) = \prod \sigma_i^2$, 
thus the constraint term in Eq.~\eqref{eq_main_func_pemalty_det}
is provided to avoid the worst case of $\prod \sigma_i^2$ being exponentially small 
(e.g., the existence of zero or linear dependent columns/rows) and big (e.g., the existence of largely scaled columns/rows).
\begin{equation}
% \begin{small}
\label{eq_main_func_pemalty_det}
%h(\mathcal{A}) 
\resizebox{.88\hsize}{!}{$
\tau(\mathcal{A})=\frac{1}{4Tk \log(k)}\sum_{t =1}^{T}\left(\log \left[\nu + \frac{1}{k} \text{det}({\mathbf{A}^{t}}^\top {\mathbf{A}^{t}}) \right] \right)^2 $}
%  h(\mathcal{A})     = \frac{1}{ 4T} \sum_{t =1}^T\left(\log \big( + \text{det}({\mathbf{A}^{t}}^\top {\mathbf{A}^{t}}) \big) \right)^2, 
%
% \end{small}
\end{equation}
with $0< \nu \ll 1$ being a small smoothing parameter.
Additionally, the penalty $\tau(\mathcal{A})$ also promotes the full rank of the elements in $\mathcal{A}$, 
i.e., $\sigma_{\min}(\mathbf{A}^{(t)}) > 0$,
as well as the full rank of $\mathcal{A}$ in matrix-vector-product, shown in 
Eq.~\eqref{eq_separable_multi_layer_tensor}.
Such two constraints $\rho({\mathcal{A}})$ and $ \tau(\mathcal{A}) $ work together to achieve a moderate condition number for the tensor system.

\subsubsection{The Objective Function for \emph{RLST} model}
\label{sec:023}
All above concepts allow us to develop learning paradigms for further promoting % construct regularization for promoting 
the \textit{lightweight} and the \textit{Robustness} by appealing to the classic concepts.
%
% By considering minimizing 
To minimize the empirical risk $\mathbb{E}_{(\mathcal{X}, {z})\sim \mathfrak{D}}[L(\mathcal{A}, \mathcal{X}, {z})]$,
%associated with the classification-driven mode $f$,  
we now construct a unified cost function for the classification model $f$ to 
jointly learn \emph{robust} and \emph{lightweight} parameters with separable structures, as 
%%%%%%%%%%%%%%%%%%%%%%%%%%%%%%%%%%%%%%%%%%%%%%%%%%%%%%%%%%%%%%%%%%%%%%%%%%%%%%%%%%%%%%%%%%%
% \begin{equation}
% 	\begin{split}
% 		\label{eq_final_cost_RLST} %  \biggl\{ 
% 		%\delta = \underset{\|\delta\|_p \leq \epsilon }{ \arg\max }  L(f(\mathcal{X}+\delta), {z})
% 		\min_{\mathcal{A}} \;  \big\{  \mathbb{E}_{(\mathcal{X}, {z})\sim \mathfrak{D}}[ & L(\mathcal{A}, \mathcal{X}, {z})] 
% 		 + \mu_{1}g(\mathcal{A})  \\
% 		 & + \mu_{2}\rho(\mathcal{A})  + \mu_{3}\tau(\mathcal{A}) 
% 		 \big\},
% 	\end{split}
% \end{equation}
%
\begin{equation}
	\begin{split}
		\label{eq_final_cost_RLST} %  \biggl\{ 
		%\delta = \underset{\|\delta\|_p \leq \epsilon }{ \arg\max }  L(f(\mathcal{X}+\delta), {z})
		\min_{\mathcal{A}} \;  \big\{  \mathbb{E}_{(\mathcal{X}, {z})\sim \mathfrak{D}}[ & L(\mathcal{A}, \mathcal{X}, {z})] + \mu_{1}\rho(\mathcal{A})  \\
		               & + \mu_{2}\tau(\mathcal{A}) + \mu_{3}g(\mathcal{A}) 
		 \big\}
	\end{split}
\end{equation}
%%%%%%%%%%%%%%%%%%%%%%%%%%%%%%%%%%%%%%%%%%%%%%%%%%%%%%%%%%%%%%%%%%%%%%%%%%%%%%%%%%%%%%%%%%%%
where three weighting factors $\mu_{1}, \mu_{2}, \mu_{3} \in \mathbb{R}^{+}$ % $\mu_{1} > 0$, $\mu_{2} > 0$ and $\mu_{3} > 0$ 
 control the influence of regularizers on the final solution.
%ensure the sparsity and the condition number reduction of separable transformations $\mathcal{A}$.
% In this work, we refer to it as the \emph{ARLST} function.
% In the rest of the paper, we call this model as Robust and Lightweight model (\emph{RL} model)
% We call this module as \emph{optimized SPARse dictionary learning for Linear Discriminative Analysis ({SpARLSTDA}) }.
% to our proposed model
In the paper, we refer to Eq.~\eqref{eq_final_cost_RLST} as Robust and Lightweight  model via Separable Transformations (\emph{RLST} model).
%

%%%%%%%%%%%%%%%%%%%%%%%%%%%%%%%%%%%%%%%%%%%%%%%%%%%%%%%%%%%%%%%%%%%
%%                           SECTION 4                           %%
%%%%%%%%%%%%%%%%%%%%%%%%%%%%%%%%%%%%%%%%%%%%%%%%%%%%%%%%%%%%%%%%%%%
%\section{Optimized dictionary learning for linear discriminative projection}
\subsection{Adversarial Training with both  lightweight and  Robustness }
\label{sec:033}
%
% With the separable transformations of tensors at hand, a corresponding robust and lightweight model (\emph{RLST}) % their several smooth measures
% was introduced in above sections,
% %in Section~\ref{sec:031} and Section~\ref{sec:032}, 
% In this subsection, we further improve the \textit{robustness} of  proposed \emph{RLST}model.
% construct a unified cost function for the proposed \emph{ARLST} model to jointly learn both \textit{robustness} and \textit{lightweight}.
%
%The condition number settings in Section~\ref{sec:0312} 
The aforementioned constraints %in Section~\ref{sec:0312} 
could reduce the sensitivity of basic linear system \eqref{eq_basic_linear_trans_MD} 
to the perturbations imposed on transformation matrices $\mathcal{A}$ and responses $\mathcal{Y}$. 
However, the multiple layers of linear transformation associated with nonlinear activations also suffer from 
hand-crafted adversarial attacks, which are implemented by adding crafted perturbations onto benign examples.
% which are often specific additive noise on input signals.
% Before presenting the \emph{ARLST} model, 
% In the following, we first briefly introduce some technical details of adversarial training for tensors, which is 
% known as an efficient learning framework against adversarially crafted attacks.
% The generability of adversarial training,  the  hand-crafted attacks,worst case
In the following, we further improve the \textit{robustness} of  proposed \emph{RLST} model by using adversarial training for tensors,
% briefly introduces some technical details of adversarial training for tensors, 
which is known as an efficient learning framework against adversarially crafted attacks.

Given an input tensor signal $\mathcal{X}\in \mathbb{R}^{I_1\times I_2\times \cdots \times I_N}$ and
corresponding label ${z}$, classification model is denoted by ${z} = f(\mathcal{A}, \mathcal{X})$.
The adversary’s goal is to find an \emph{adversarial example} $\mathcal{X}^{adv} = \mathcal{X} + \delta$ that fools the classifier $f$, 
i.e., taking $\mathcal{X} + \delta$ as an input result in a much worse classification result of $f$.
Therein, $\delta$ denotes  a small additive perturbation.
Formally, the tensor adversarial examples could be generated by the solving the following maximization problem: %a perturbation $\delta$ is learned as follows
%an adversary could create a perturbation $\delta$ as follows
\begin{equation}
	\mathcal{X}^{adv} = \underset{\|\mathcal{X}^{adv} - \mathcal{X} \|_p \leq \epsilon }{ \arg\max }  L(\mathcal{A}, \mathcal{X}^{adv}, {z}),
\end{equation}
% %
% \begin{equation}
% 	\label{eq_adv_attack_MD}
% 	%\delta = \underset{\|\delta\|_p \leq \epsilon }{ \arg\max }  L(f(\mathcal{X}+\delta), {z})
% 	\mathcal{X}^{adv} = \underset{\|\mathcal{X}^{adv} - \mathcal{X} \|_p \leq \epsilon }{ \arg\max }  L(\mathcal{A}, \mathcal{X}^{adv}, {z})
% \end{equation}
% %
where $\epsilon \in \mathbb{R}^{+}$ is a small scaling parameter, % measure the loss of classifier $f$,
$L$ is the loss function used to train the adversarial model, e.g., cross-entropy for a neural network \cite{madry2017towards_iclr18},
and  $\|\cdot\|_p$ with $p \geq 0$ denotes the $\ell_p$ norm of a tensor, which is used to measure the distance between $\mathcal{X}^{adv}$ and $\mathcal{X}$.
%and  $\|\cdot\|_p$ with $p \geq 0$ denotes the $\ell_p$ norm of a tensor,
%%, defined by
%%%
%%$$\|\mathcal{X}\|_p = \left( \sum_{i_1}^{I_1} \sum_{i_2}^{I_2} \cdots \sum_{i_N}^{I_N} X_{i_1 i_2\cdots i_N}^p \right)^{1/p},$$
%%%
%which is used to measure the distance between $\mathcal{X}^{adv}$ and $\mathcal{X}$.
% Therein, $X_{i_1 i_2\cdots i_N}$ is the $(i_1 i_2\cdots i_N)-$element of $\mathcal{X}$.
%
% According to the adversarial robustness literatures of 
% %\cite{madry2017towards_iclr18,Ye2019Adversarial_both_ICCV,tramer2018ensemble_goodfellow_iclr,dhillon2018stochastic_iclr,tramer2019adversarial_nips,cARLSTini2019evaluating_arxiv}, dhillon2018stochastic_iclr
% \cite{madry2017towards_iclr18, Ye2019Adversarial_both_ICCV, tramer2018ensemble_goodfellow_iclr, 
% 	tramer2019adversarial_nips, carlini2019evaluating_arxiv},
The goal of tensor adversarial training %\cite{madry2017towards_iclr18} % \emph{adversarial robustness} 
against adversarial examples $\mathcal{X}^{adv}$ 
is to find model parameters $\mathcal{A}$ via optimizing the following min-max Empirical risk problem
\begin{equation}
	\label{eq_adv_robustness_MD}
	%\delta = \underset{\|\delta\|_p \leq \epsilon }{ \arg\max }  L(f(\mathcal{X}+\delta), {z})
	\min_{\mathcal{A}} \; { \mathbb{E}_{(\mathcal{X}, {z})\sim \mathfrak{D}}\underset{\|\mathcal{X}^{adv} - \mathcal{X} \|_p \leq \epsilon }{ \arg\max }  L(\mathcal{A}, \mathcal{X}^{adv}, {z}) },
\end{equation}
where $\mathfrak{D}$ denotes the underlying training data distribution.
% Therein, the inner maximization problem aims to find adversarial attack examples to maximally mislead the classification model $f$, 
% and the outer minimization problem aims to retrain the model parameters to minimize the loss caused by these attack examples.
% % corresponds to training over these attack examples.
%
%%%%%%%%%%%%%%%%%%%%%%%%%%%%%%%%%%%%%%%%%% 
%-------------------------------------------------------------------------
\begin{algorithm}[t]
	\caption{:Algorithm for \emph{ARLST} function}
	\label{alg_aRLST}
	\begin{algorithmic}[1] 
		\STATE \textbf{Input:} 
		%     Initial dictionary $\mathbf{D}^{(0)} \in \mathcal{S}(m,k)$, 
		Training set $\mathcal{D}$, iterations $K$, model $f_{\mathcal{A}}$ 
		with separable parameters $\mathcal{A}$,  batch size $T$,
		$\mu_{1}, \mu_{2}, \mu_{3}$, attacks
		%		$\lambda_1$,  $\lambda_2$, $\lambda_3$,
		%		thresh ${\bm{\epsilon}}_1, {\bm{\epsilon}}_2$, $T$. 
		%
		\STATE \textbf{Output:} parameters $\mathcal{A}$
		\FOR {$ i \;= \; 1 \; to\; \textit{K}$}
		\STATE Read mini-batch $\mathcal{X} \subset \mathcal{D}$ 
		\STATE Given $\mathcal{A}$, generate $T$ adversarial examples 
		$\mathcal{X}^{adv} := \{\mathcal{X}^{adv}_{1},\cdots, \mathcal{X}^{adv}_{T}\}$ 
		from corresponding clean signals $\mathcal{X} $
		by solving the inner max in Eq.~\eqref{eq_final_cost_ARLST}
		\STATE Apply Adam optimizer on the outer min in Eq.~\eqref{eq_final_cost_ARLST}, 
		to update $\mathcal{A}$ by using learned $\mathcal{X}^{adv}$
		\ENDFOR
%			\vspace*{-4mm}
	\end{algorithmic}
\end{algorithm}
%%%%%%%%%%%%%%%%%%%%%%%%%%%%%%%%%%%%%%%%%% 
%
%%%%%%%%%%%%%%%%%%%%%%%%%%%%%%%%%%%%%%%%%%%%%%%%%%%%%%%%%%%%%%%%%%%%
%%%                           SECTION 4.3                           %%
%%%%%%%%%%%%%%%%%%%%%%%%%%%%%%%%%%%%%%%%%%%%%%%%%%%%%%%%%%%%%%%%%%%%
%\subsubsection{The Objective Function for \emph{ARLST} model}
%%
%\label{sec:043}
%%
% The goal of \emph{RLST} is to train a robust and lightweight model from the linear nature of the deep learning model.
% %
% As introduced in Section~\ref{sec:032}, the linear transformation with separable structured parameters associated with the sparsity promotion 
% reduces the computational load of model.
% Furthermore, the settings for condition number within the framework of adversarial training associated with min-max optimization could exactly 
% % the min-max optimization based adversarial training framework could exactly 
% improve the robustness against various perturbations. 

%To summarize, we construct the following cost function to jointly learn both a 
%sparsifying dictionary and an orthogonal transformation, i.e.,
By combining the regularizers discussed above, we construct the following cost function  
to jointly learn robust and lightweight parameters with separable structures, as 
%jointly learn both the dictionary and the linear transition matrix, i.e.,
%
\begin{equation}
	\begin{split}
		\label{eq_final_cost_ARLST} % \biggr\}
		%\delta = \underset{\|\delta\|_p \leq \epsilon }{ \arg\max }  L(f(\mathcal{X}+\delta), {z})
		\min_{\mathcal{A}} \;  \big\{  & \mathbb{E}_{(\mathcal{X}, {z})\sim \mathfrak{D}}    \underset{\|\mathcal{X}^{adv} - \mathcal{X} \|_p \leq \epsilon }{ \arg\max }  L(\mathcal{A}, \mathcal{X}^{adv}, {z})   \\
		&  + \mu_{1}\rho(\mathcal{A})+\mu_{2}\tau(\mathcal{A})+\mu_{3}g(\mathcal{A}) 
		\big\}
	\end{split}
\end{equation}
%+ \mu_{1}g(\mathcal{A})
with $\mu_{1}, \mu_{2}, \mu_{3}\in \mathbb{R}^{+}$.
%the three weighting factors $\mu_{1}, \mu_{2}, \mu_{3} \in \mathbb{R}^{+}$. % $\mu_{1} > 0$, $\mu_{2} > 0$ and $\mu_{3} > 0$ 
%control the influence of the three regularizers on the final solution.
%
In this work, we refer to it as the Adversarial \emph{RLST} (\emph{ARLST}) function.

\paragraph{The gradients}
The key requirement for developing a gradient algorithm to optimize the cost function Eq.~\eqref{eq_final_cost_RLST} or Eq.~\eqref{eq_final_cost_ARLST}, 
 is the differentiability of \textit{LRST/ALRST} function w.r.t. $\mathcal{A}$.
By taking \textit{ALRST} as an example, let ${J}$ and $\mathbf{A}$ denote by the cost function in Eq. \eqref{eq_final_cost_ARLST} and one of its separable parameter,
the Euclidean gradient $\nabla{J}(\mathbf{A})$ is computed as
\begin{equation}
    \begin{split}
    \label{eq_final_cost_ALRST_gradient}
    \nabla{J}(\mathbf{A}) = &\nabla_L(\mathbf{A}) + \mu_1 \nabla_{\rho} (\mathbf{A}) + \mu_2 \nabla_{\tau}(\mathbf{A}) \\
    &+ \mu_3 \nabla_g(\mathbf{A}). 
    \end{split}
\end{equation}

Since the optimization on DNNs is nowadays well established, 
% the differentiability of the function w.r.t. $\bm{\theta}_{\text{DNN}}$ 
$\nabla_L(\mathbf{A}) $ is commonly available and its Euclidean gradient
depends on the DNN architectures of use \cite{he2016deep_cvpr_resnet, hochreiter1997long_nc}.   % at disposal  determined
In Eq.~\eqref{eq_final_cost_ALRST_gradient}, it has
\begin{equation}
\begin{aligned}
\nabla_{{\rho}}(\mathbf{A}) & = \dfrac{1}{k^2}\mathbf{A},\\
\nabla_{\tau}(\mathbf{A}) & = \dfrac{\eta}{ \log(k) } \mathbf{A} (\eta \mathbf{A}^\top \mathbf{A})^{-1}.
%      \nabla_{{\rho}}(\mathbf{A}) & = \dfrac{1}{k} \sum_{i = 1}^{k}\dfrac{1}{p} \sum_{j = 1}^{k} (a_{ij}^2 + 1)^{\dfrac{p}{2}} 
%      \sum_{j=1}^{k}( a_{ij} (a_{ij}^2 + 1)^{\dfrac{p}{2}-1} \mathbf{E}_{ij} ), \\ 
\end{aligned}
\end{equation}
%%

%  weadmit this observed phenomena, and 
Here, we enforce the sparsity of $\mathbf{A}$ 
by minimizing a $\ell_p$ norm with $0\leq p \leq 1$.
In practice, we use the following penalty term to constrain $\mathbf{A}$ instead of Eq.~\eqref{eq_sparsity} , 
\begin{equation}
\label{eq_main_func_sparsity}
g(\mathbf{A}) = \frac{1}{2k_1} \sum_{i = 1}^{k_1} (\frac{1}{p} \|\mathbf{a}_{i,:}\|_p^p )^2,
\end{equation}
% $0\leq p \leq 1$ and
with $\mathbf{a}_{i,:}$ being the $i^{\text{th}}$ row of $\mathbf{A}$. 
However, it is known that above $\ell_p$ norm is non-smooth. In order to make the global 
cost function differentiable, 
we exchange Eq.~\eqref{eq_main_func_sparsity} with a smooth approximation that is concretely given as
\begin{equation}
\label{eq_main_func_sparsity_appro}
g(\mathbf{A}) = \frac{1}{2k_1} \sum_{i = 1}^{k_1} ( \sum_{j = 1}^{k_2}(\mathbf{a}_{ij}^{2}+  \varpi)^{{p}/{2}} )^2 ,
\end{equation}
with $0 < \varpi < 1$ being a smoothing parameter. Then, the Euclidean gradient $\nabla_{g}(\mathbf{A})$ is computed as
%%
%\begin{equation}
%\nabla_{\widetilde{J}}(\mathbf{A}) = \sum\limits_{i=1}^{n} \dfrac{1}{n} {\bm{\phi}}_{{\mathbf{x}}_i}  \mathbf{e}_{i}^{\top} 
%+\mu_1 \nabla_{\rho}(\mathbf{A}) 
%%

%\end{equation}
%%%
%with
%

\begin{equation}
\begin{footnotesize}
\begin{aligned}
\nabla_{g}(\mathbf{A})= \dfrac{1}{k_1} \sum_{i = 1}^{k_1}\dfrac{1}{p} \sum_{j = 1}^{k_2} (\mathbf{a}_{ij}^2 + \varpi)^{\dfrac{p}{2}} 
\sum_{j=1}^{k_2}( \mathbf{a}_{ij} (\mathbf{a}_{ij}^2 + \varpi)^{\dfrac{p}{2}-1} \mathbf{E}_{ij} )   
%       \nabla_{{\rho}}(\mathbf{A}) & = \dfrac{\eta}{k \log(k) } \mathbf{A} (\eta \mathbf{A}^\top \mathbf{A})^{-1}.
\end{aligned}
\end{footnotesize}
\end{equation}
 By $\mathbf{E}_{ij}$, we denote a matrix whose $i^{th}$ entry in the 
$j^{th}$ column is equal to one, and all the rests are zero. 

With the differentiability of the \emph{RLST/ARLST} function,
smooth solvers, such as stochastic gradient descent algorithms, can be used for optimization.
%as shown in Appendix~\ref{sec:appendix_opti}. 
A generic framework of our \emph{ARLST} algorithm is summarized in Algorithm~\ref{alg_aRLST}.

\begin{table*}[h!]\small
	\centering
	\caption{Comparison between a MLP and its variants.}
	\begin{tabular}{cccccccc}
		\hline
		  & & MLP& MLP-HYDRA & MLP-ADMM& MLP-ATMC& MLP-RLST  \\ \hline
		\multirow{2}{*}{MNIST} & Parameters & \textup{322M} & \textup{5.25M}($61\times$) &  \textup{5.25M}($61\times$) & \textup{5.25M}($61\times$) & \textbf{5.25M}($61\times$) \\ 
		& Accuracy & \textbf{97.99\%}  & 74.48\% &  92.90\% & 95.83\% & \textbf{97.99\%} \\ \hline
		\multirow{2}{*}{CIFAR-$10$} & Parameters    & \textup{834.5M}  & \textup{14.5M}($57\times$)   & \textup{14.5M}($57\times$)   & \textup{14.5M}($57\times$) & \textbf{14.5M}($57\times$)    \\ 
		& Accuracy &  56.42\%  & 47.54\%  & 37.97\%  & 52.42\% & \textbf{67.82\%} \\ \hline
	\end{tabular}
	\label{table_autoencoder}
\end{table*}

%------------------------------------------------------------------------

\section{Experimental Evaluations}
\label{sec:04}
In this section, we investigate the performance of the proposed \emph{RLST/ARLST} function 
from two aspects: the size of model parameters and the model \textit{robustness} against different perturbations.
% we will  experiments on synthesized the CIFAR-$10$ datasets  to testify the effectiveness of the proposed ARLST
% under different perturbations.
%
For the convenience of referencing, we adopt the following way to 
name the algorithms for comparison: for example, the VGG-$16$ \cite{simonyan2014very_vgg16} under the framework of the \emph{ARLST} 
algorithm is referred to as the VGG$16$-\emph{ARLST}.

\begin{table*}[h!]\small
    \centering
    \caption{\label{table_adv_cifar} 
 		Comparison between a VGG-16 and its variants under adversarial training with PGD attack.
 	    	}
    \begin{tabular}{cccccc}
    \hline
                      & & VGG16-HYDRA & VGG16-ADMM & VGG16-ATMC & VGG16-ARLST    \\ \hline
    & \textbf{Cr} & \textbf{NA}/\textbf{RA} (\textbf{Var}) & \textbf{NA}/\textbf{RA} (\textbf{Var})& \textbf{NA}/\textbf{RA} (\textbf{Var}) & \textbf{NA}/\textbf{RA} (\textbf{Var})\\ \hline
    \multirow{6}{*}{\begin{tabular}{@{}c@{}} \rotatebox[origin=c]{90}{CIFAR-10} \end{tabular}} &$1 \times $(pretrain, \textbf{NAT}) & $96.1/3.5$ &$96.1/3.5$&$96.1/3.5$& $96.1/3.5$    \\    
    &$1 \times $(pretrain, \textbf{AT}) & $77.7/47.7(0.52/0.13)$ &$76.0/44.8(0.25/0.16)$&$76.9/45.5(0.21/0.11)$&  $\textbf{77.8/47.7}(0.29/0.10)$    \\ 
    &$10 \times $          & $75.7/46.2(0.25/0.16)$ &$75.9/44.8(0.20/0.36)$    
         &$75.6/44.8(0.18/0.19)$&$\textbf{76.6/46.3}(0.18/0.03)$   \\ 
    &$20 \times $          & $74.5/44.9(0.15/0.05)$ &$74.7/43.7(0.15/0.06)$  
        &$73.9/42.9(0.13/0.20)$&$\textbf{75.8/45.6}(0.11/0.01)$     \\ 
    &$100 \times $         & $69.8/40.2(0.09/0.26)$ & $68.8/40.4(0.20/0.63)$  
        &$69.8/41.1(0.12/0.19)$&$\textbf{70.0/41.3}(0.05/0.15) $  \\
    &$200 \times $        & $62.4/35.5(0.23/0.16)$ &$60.6/30.5(0.05/0.32)$&$64.0/36.1(0.19/0.35)$ 
         &$\textbf{64.3/37.5}(0.02/0.09)$   \\ \hline
     \multirow{6}{*}{\begin{tabular}{@{}c@{}} \rotatebox[origin=c]{90}{SVHN} \end{tabular}}&$1 \times$(pretrain, \textbf{NAT})&$96.2/1.2$& $96.2/1.2$& $96.2/1.2$& $96.2/1.2$ \\ 
    &$1 \times $(pretrain, \textbf{AT}) & $90.5/53.5(0.35/0.23)$& $89.7/52.3(0.20/0.09)$& $90.2/52.2(0.23/0.06)$&  $\textbf{90.6/53.9}(0.10/0.01)$     \\ 
    &$10 \times $          & $89.2/52.4(0.20/0.29)$& $88.2/51.6(0.32/0.19)$&$88.6/51.7(0.32/0.15)$& 
     $\textbf{89.6/52.9}(0.08/0.05)$  \\ 
    &$20 \times $          & $85.5/51.7(0.23/0.32)$& $85.1/50.9(0.20/0.16)$&$85.4/50.8(0.11/0.20)$& 
     $\textbf{87.5/52.8}(0.11/0.12)$  \\ 
    &$100 \times $          & $84.3/46.8(0.22/0.19)$& $83.9/45.6(0.32/0.09)$&$80.5/46.8(0.25/0.10)$&  
     $\textbf{84.6/51.6}(0.23/0.05)$\\ 
    &$200 \times $          & $32.9/23.1(0.10/0.26)$& $30.1/22.5(0.12/0.32)$&$70.0/30.7(0.10/0.20)$& 
     $\textbf{73.1/34.7}(0.05/0.10)$  \\ \hline
    \end{tabular}
\end{table*}
\paragraph{Datasets.}
Our method is validated by experiments on image datasets, such as SVHN \cite{netzer2011reading},Yale-B, MNIST,
CIFAR-$10$ \cite{krizhevsky2009learning_cifar10}, CIFAR-$100$ \cite{krizhevsky2009learning_cifar10} 
and 
ImageNet (ILSVRC2012) \cite{ILSVRC15_ijcv}. 
%
% The Mini-ImageNet is manually selected from ImageNet, with $10$ categories and $13,000$ images.
%
%Moreover, we report the model performance by the wildly-used metrics of robust accurary: It refers to the percentage of correctly classified adversarial examples generated with PGD \cite{madry2017towards_iclr18} attack or {\color{red} FGSM \cite{goodfellow2014explaining_iclr} attack}.

% The PGD attack is used 
\paragraph{Implementation Settings.}
We report the model performance on \textit{robustness} under 
the FGSM attack \cite{goodfellow2014explaining_iclr},
the PGD attack \cite{madry2017towards_iclr18}, 
the AutoAttack (AA) \cite{croce2020reliable}
and the Square Attack (SA) \cite{andriushchenko2020square}.
The PGD attack is known as the strong first-order attack.
% and if the model demonstrates high robustness under the PGD attack, 
% then the model has resistance to other first-order attacks \cite{madry2017towards_iclr18}. 
The AA is an ensemble of complementary attacks which combine new versions of PGD with some other attacks.
%with FAB and some other attacks.
The SA is a black-box attack, which is an adversary attack without any internal knowledge of the targeted network.
Unless otherwise specified, 
we set the PGD attack and the AA attack with the perturbation magnitude $\epsilon = 0.031$, 
the attack iteration numbers $n = 10$, and the attack step size $s = 0.0078$.
We also set the number of queries in SA as 100,
the perturbation magnitude of FGSM $\epsilon = 0.015$.

%

%Moreover, we report the model performance by the wildly-used metrics of robust
%accurary (in the next subsection, I will introduce imoportant metrices in detail).
%
We evaluate the model performance by using the following metrics, as
%
%\end{enumerate}
\romannumeral 1) \textbf{Compression ratio (\textbf{Cr})}: the division of regular model size by lightweight model size;
% the division of original model size by model size after the \textit{lightweight};
% regular model size
\romannumeral 2) \textbf{Natural Accuracy (NA)}: the accuracy on classifying benign images;
%It is the percentage of correctly classified benign (non-modified) images.
\romannumeral 3) \textbf{Robustness Accuracy (RA)}:  the accuracy on classifying images corrupted by adversarial attack;
%It refers to the percentage of correctly classified adversarial examples generated with PGD \cite{madry2017towards_iclr18} attack.
\romannumeral 4) \textbf{Var}: the variance of accuracy results from multiple experiments.
% If no variance is provided, it means that the data comes from the paper.
%\romannumeral 5) \textbf{$\Delta$}: refers to how much our results is greater than the comparisons;
%\romannumeral 5) \textbf{AT} denotes the Adversarial Training defined in %\cite{madry2017towards_iclr18} and \cite{goodfellow2014explaining_iclr}
%      and \textbf{NAT} denotes the results under natural training, i.e., 
%     evaluating a learning model without adversarial training with respect to specific known attack.
In addition, \textbf{AT} denotes the Adversarial Training method defined in \cite{zhang2019theoretically_icml} and 
% \cite{madry2017towards_iclr18} and \cite{goodfellow2014explaining_iclr},  
\textbf{NAT} denotes the results under natural training, i.e., 
evaluating a learning model without adversarial training with respect to specific known attack.
% This refers to improvement of our method.

%
% Then, \cite{lin2019defensive_iclr} proposed a novel defensive quantization
% method by controlling the neural network’s Lipschitz constant during quantization.
% And then, the Defensive Quantization method in \cite{lin2019defensive_iclr} is an attractive way to optimize the robustness and efficiency in deep learning models. This approach mitigates the adversarial attacks during quantization, whereas our work focuses on the pruning methodology. Although we didn't compare our method with \cite{lin2019defensive_iclr} temporarily, we will add the quantization method after the pruning phase in our future work. 
% And then, the Defensive Quantization method in \cite{lin2019defensive_iclr} is an attractive way to optimize the robustness and efficiency in deep learning models. 
% This approach mitigates the adversarial attacks during quantization, 
% \cite{goldblum2019adversarially_distillation} proposed an Adversarially Robust Distillation, which leveraging knowledge distillation to produce high robustness student network. 

% Weight sharing and quantization methods assume that many weights have similar values, and can thus be grouped in order to reduce the number of free parameters.

%
Three well-known and most related methods, ADMM \cite{Ye2019Adversarial_both_ICCV}, ATMC \cite{gui2019model_nips}, and HYDRA \cite{sehwag2020pruning_adversarial},
are used for the comparison.
All networks are trained with 100 epochs for all experiments in this paper, and they are conducted on NVIDIA RTX $2080Ti$ GPU ($10$ GB memory for each GPU).
\textbf{Unless otherwise specified, the results of all tables are presented in percent. }

%%%%%%%%%%%%%%%%%%%%%%%%%%%%%%%%%%%%%%%%%%%%%%%%%%%%%%%%%%%%%%%%
%In the experiment, we selected HYDRA \cite{sehwag2020pruning_adversarial} as the
%benchmark algorithms.
%%HYDRA是一种新的剪枝技术，它通过将鲁棒训练制定为经验风险最小化问题来意识到鲁棒训练目标。他们采用了基于重要性分数的优化技术，并对重要性分数进行了缩放初始化。然后在对抗训练过程中，不停地更新重要性分数，最后根据重要性分数来决定哪些连接需要被剪除
%HYDRA \cite{sehwag2020pruning_adversarial}is a novel pruning technique, which is aware of the robust training objective by formulating robust training as an empirical risk minimization problem.
%They employ an importance score based optimization technique with scaled initialization of importance scores.
%During the adversarial training process, the importance scores are constantly updated, and finally the importance scores are used to determine which connections need to be pruned.
%Note that the two methods use the same adversarial training framework(Trades-off\cite{zhang2019theoretically_icml}). 
%The related codes for them are released by the authors in GitHub. 
%We follow their basic settings without much post processing.

%
%%%%%%%%%%%%%%%%%%%%%%%%%%%
%%%%%%%%%%%%%%%%%%%%%%%%%%%%%%%%%%%%%%%%%
\begin{figure}[h!]
	\centering
	\includegraphics[width=0.7\linewidth]{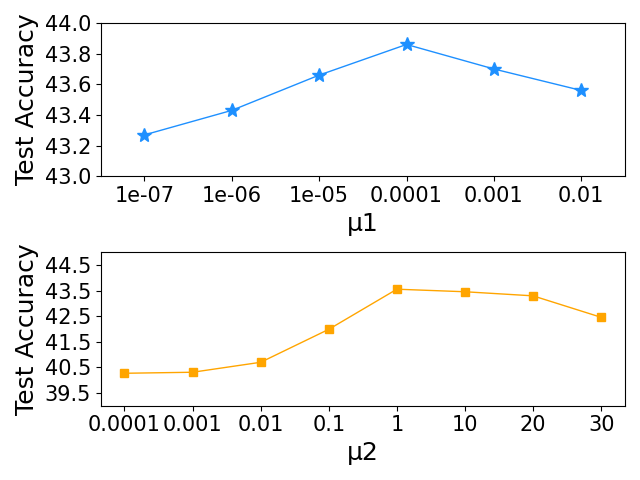}
	\caption{ Impact of condition number regularization operator on test accuracy. 
	%on the USPS digits
% 	The horizontal axis represents different regularization coefficients, while the vertical axis represents training accuracy under  different  coefficients.
	}
	\label{fig:xiaorong}
		\vspace*{-4mm}
\end{figure}
% %%%%%%%%%%%%%%%%%%%%%%%%%%%%
\begin{figure}[h!]
	\centering
	\includegraphics[width=0.7\linewidth]{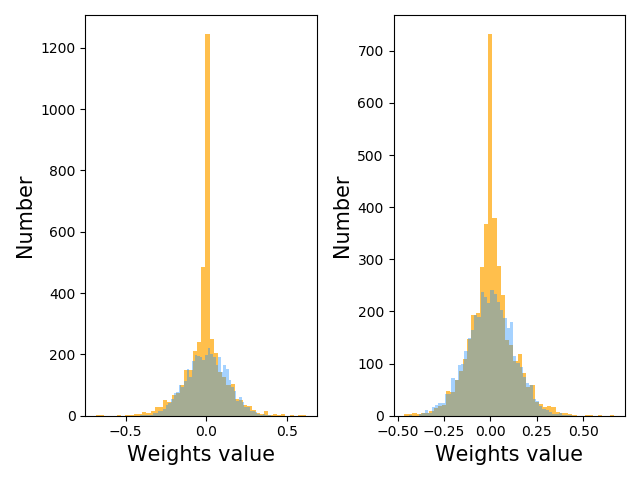}
	\caption{ % The distribution of two separable weighted matrices.
	% separable weights of two separated matrices. 
	Distribution of parameters of two separable matrices with (yellow part) and without (dark part) sparse regularization.}
	\label{fig: A11}
		\vspace*{-4mm}
\end{figure}
% %
% \begin{figure}[!ht]
% 	\centering
% 	\subfigure[ ResNet-$5$: $k(\A \otimes \B)$]{
% 		\label{fig:regu1}
% 		\includegraphics[width=0.225\textwidth]%[width=0.3\textwidth,height=0.2\textwidth]
% 		{fig/kw_original.jpeg}
% 	}
% 	%\hspace{3mm}
% 	\subfigure[ ResNet-$5$: $k(\A)$ and $k(\B)$]{
% 		\label{fig:regu2}
% 		\includegraphics[width=0.225\textwidth]
% 		{fig/kAB_original.jpeg}
% 	}
% 	%	\hspace{3mm}
% 	\subfigure[ ARLST-$5$: $k(\A \otimes \B)$ ]{
% 		\label{fig:regu3}
% 		\includegraphics[width=0.225\textwidth]
% 		{fig/kw_ARL.jpeg}
% 	}
% 	% \hspace{3mm}
% 	\subfigure[ARLST-$5$: $k(\A)$ and $k(\B)$ ]{
% 		\label{fig:regu4}
% 		\includegraphics[width=0.225\textwidth]
% 		{fig/kAB_ARL.jpeg}
% 	}
% 	\vspace{-1mm}
% 	\caption{
% 		\label{fig_condition_number2}
% 		%
% 		Condition numbers curve following the increase of training iterations.
% 		We set $\mu_{1} = 0.5$, $\mu_{2} = 1$, and $\mu_{3} = 0.1$.
% 		%
% 		The final conditional numbers for ResNet-$5$: $k(\A \otimes \B) = 1546.4416$, $k(\A ) = 1546$,  $k(\B ) = 32.9219$, $acc = 46.9730\%$.
% 		The final conditional numbers for ARLST-$5$: $k(\A \otimes \B) = 14.4577$, $k(\A ) = 2.2819$,  $k(\B ) = 6.4452$, $acc = 97.58\%$.
% 	}
% \end{figure}
% %\vspace{-1.5mm}
%%%%%%%%%%%%%%%%%%%%%%%%%%%%
%%%%%%%%%%%%%%%%%%%%%%%%%%%%%%%%%%%%%%%%%%%%%%%%%%%%%%%%%%%%%%%%%%%%%%%%%%%%%%%%%%%%%%%%%%%%%%%%%%%%%%%%             Weighing the Regularisers         
%%%%%%%%%%%%%%%%%%%%%%%%%%%%%%%%%%%%%%%%%%%%%%%%%%%%%%%%%%%%%%%%%%%%%%%%%%%%%%%%%%

% The PGD attack is used 
\paragraph{Weighing the Regularisers.}
Training an \emph{ARLST} model is usually computationally expensive.
%
% We firstly investigate the impact of various factors of the learning model.
% All experiments in this subsection were conducted on the CIFAR-$10$ dataset.
%
Here, we investigate the impact of the three regularizers $\rho$, $\tau$ and $g$
on the performance of the \emph{ARLST} under the PGD attack, on the CIFAR-$10$ dataset, 
i.e., the inference of weighing parameters
$\mu_{1}$, $\mu_{2}$ and $\mu_{3}$ in Eq.~\eqref{eq_final_cost_RLST} and Eq.~\eqref{eq_final_cost_ARLST}.

We first investigate the impact of $\mu_1$ and $\mu_2$, i.e., $\rho$ and $\tau$, as shown in the Fig.~\ref{fig:xiaorong}.
We set $\mu_2 = 0$, $\mu_3 = 0$ when $\rho$ is investigated, and $\mu_1 = 0$, $\mu_3 = 0$ for investigating $\tau$.
Experimental results show that suitable choices of $\mu_1$ and $\mu_2$ can improve the performance of the \emph{ARLST} method.
$\mu_{1} = 0.0001$ and $\mu_{2} = 1$ achieve the superior performance.

Similarly, Fig.~\ref{fig: A11} 
% plots the distribution of parameters with (yellow part) and without (dark part) $g$. 
% The figure 
shows that the regularizer $g$ can make the parameters more sparse and improve the 
\textit{lightweight} of the model.

\subsection{MLPs}
\label{sec:041}

In order to initially evaluate the performance on improving the model \textit{lightweight} by \emph{RLST},
a simple MLP with three layers is tested on MNIST and CIFAR-$10$,
in the framework of \emph{RLST}, with $T = 2$, namely, MLP-\emph{RLST}. 
For comparison, the MLP is also compressed by using
ADMM, ATMC and HYDRA
% ADMM \cite{Ye2019Adversarial_both_ICCV}, ATMC \cite{gui2019model_nips}, and HYDRA \cite{sehwag2020pruning_adversarial} 
with the setting of \textbf{NAT},
namely, MLP-ADMM, MLP-ATMC and MLP-HYDRA.

%We build the Multilayer perceotrons by using the original FC layers and the Separable Structured Transformations respectively, namely MLP and MLP-RLST. 
%Then train with MNIST \footnote[1]{\url{http://yann.lecun.com/exdb/mnist/}} and CIFAR-$10$ \cite{krizhevsky2009learning_cifar10}. 

As shown in Table.~\ref{table_autoencoder}, MLP-\emph{RLST} can reduce the model parameters of a MLP by more than $61\times$ on MNIST,
and more than $57\times$ on CIFAR-$10$, while ensuring the same even better classification results in comparison of MLP. 
These results are much better than other methods, with the similar compression ratio.
This indicates that  MLP-\emph{RLST} has more power on feature expressiveness  when encountering a very high compression ratio.
It is worth noting that on the CIFAR-10 dataset, MLP-RLST even outperforms the accuracy of MLP, which may be due to the fact that the former has less risk of overfitting.

%%%%%%%%%%%%%%%%%%%%%%%%%%%%%%%%%%%%%%%%%%%%%%%%%%%%%%%%%%%%%%%%%%%%%%%%%%%%%%%%%%%%%%%%%%%%%%%%%%%%%%%%%%%%%%%%%     ARLST with Adversarial Training       %%%%%%%%%%%%%%%%%%%%%%%%
%%%%%%%%%%%%%%%%%%%%%%%%%%%%%%%%%%%%%%%%%%%%%%%%%%%%%%%%%%%%%%%%%%%%%%%%%%%%%%%%%%%%%%%%%%
\begin{table*}[h!]\small
    \centering
    \setlength{\tabcolsep}{1mm}{
	\caption{\label{table_vit}
	Comparison between a Vision Transformer and its variants under adversarial training with FGSM attack.}
    \begin{tabular}{cccccc}
    \hline
     &  & ViT-HYDRA & ViT-ADMM & ViT-ATMC & ViT-ARLST \\ \hline
     & \textbf{Cr} & \textbf{NA}/\textbf{RA} (\textbf{Var}) & \textbf{NA}/\textbf{RA} (\textbf{Var})& \textbf{NA}/\textbf{RA} (\textbf{Var}) & \textbf{NA}/\textbf{RA} (\textbf{Var})\\ \hline
    \multirow{3}{*}{\begin{tabular}{@{}c@{}} \rotatebox[origin=c]{90}{MNIST} \end{tabular}} & $1\times$(pretrain,\textbf{NAT})
    &$99.47/6.41(0.32/0.20)$ &$99.47/6.41(0.32/0.20)$ &$99.47/6.41(0.32/0.20)$ &$99.47/6.41(0.32/0.20)$ \\ 
    &$1\times$(pretrain,\textbf{AT}) &$97.32/89.86(0.25/0.22)$ &$96.98/89.21(0.40/0.19)$ &$97.58/90.12(0.15/0.09)$ &$\textbf{97.98/90.94}(0.20/0.10)$ \\ 
    &$3\times$(\textbf{AT}) &$92.60/81.12(0.31/0.23)$ &$91.98/81.08(0.23/0.31)$ &$94.30/83.34(0.25/0.16)$ &$\textbf{94.65/83.96}(0.21/0.15)$ \\ \hline
    \multirow{3}{*}{\begin{tabular}{@{}c@{}} \rotatebox[origin=c]{90}{Yale-B} \end{tabular}} &  $1\times$(pretrain,\textbf{NAT})
    &$99.74/0.79(0.33/0.15)$
    &$99.74/0.79(0.33/0.15)$
    &$99.74/0.79(0.33/0.15)$ &$99.74/0.79(0.33/0.15)$\\ 
    & $1\times$(pretrain,\textbf{AT})
    &$95.96/90.88(0.31/0.23)$ 
    &$95.20/90.10(0.19/0.22)$ 
    &$96.50/91.03(0.29/0.20)$ &$\textbf{96.65/92.93}(0.10/0.25)$  \\ 
    &$3\times$(\textbf{AT}) 
    &$93.68/89.86(0.10/0.20)$ &$93.35/89.05(0.25/0.32)$ &$94.65/90.89(0.23/0.53)$ &$\textbf{95.83/92.74}(0.15/0.13)$ \\ \hline
    \end{tabular}}
\end{table*}
\subsection{ARLST with Adversarial Training}
\label{sec:042}

%\subsection{Comparison on classification with SOTA models}
%\subsection{Comparison on Image Classification}
%
In order to evaluate both the \textit{robustness} and the \textit{lightweight} of proposed \emph{ARLST} model, 
the \textbf{NA}, the \textbf{RA},
% the natural accuracy (NA), the robustness accuracy (RA), 
 as well as the amount of model parameters, %with and without Adversarial Training
are considered for comparison with SOTA methods, ATMC, %\cite{gui2019model_nips}, 
ADMM %\cite{Ye2019Adversarial_both_ICCV} 
and HYDRA. %\cite{sehwag2020pruning_adversarial}.
We follow their basic settings without much post processing. 
By taking VGG-$16$ performed on CIFAR-$10$ dataset as an example,
we follow its basic architecture and construct  VGG-$16$-\emph{ARLST} network.
%
% wThe results generated by ATMC, ADMM and HYDRA showed in this section are reported in references.
% Note that, the related codes for them are released by the authors in GitHub.

As shown in Table.~\ref{table_adv_cifar},
%and Table~\ref{table_adv_svhn},
% and  Table~\ref{table_adv_imagenet}, 
by using the same pre-trained setting, VGG-$16$-\emph{ARLST} outperforms ATMC, ADMM and HYDRA
in two datasets under different levels of compression ratio.
It is worth noting that the VGG-$16$-\emph{ARLST} achieves overwhelming advantage when the compression ratio is very high, e.g., $200$ times. Additionally, the VGG-$16$-\emph{ARLST} has a smaller variance of the accuracies for five runs in the experiments.
These results suggest that the VGG-$16$-\emph{ARLST} has higher stability than above SOTA methods. 

By taking the experimental results on SVHN as an example, 
% when the compression ratio is $90\%$, $95\%$, $99\%$ and $99.5\%$, 
the \textbf{RA} of the  VGG-$16$-\emph{ARLST} achieves 
% $0.13\%$, $0.67\%$, $1.08\%$ and $6.48\%$ 
% $2.7\%$, $3.78\%$, $4.3\%$, $7.56\%$ and $7.8\%$
$0.4\%$, $0.5\%$, $1.1\%$, $4.8\%$ and $11.6\%$
higher than that of HYDRA, 
with the compression ratio as $1\times$, $10\times$, $20\times$, $100\times$ and $200\times$, respectively.
It is apparent that the VGG-$16$-\emph{ARLST} achieves more advantages as the compression ratio increases.
%随着压缩比的增大，ARLST在鲁棒性精度上的表现也越来越好。
% The improvement effect of the VGG-$16$-\emph{ARLST} model becomes more and more significant as the compression ratio increases.
This advantage may be traced to the joint optimization of the pruning and the condition number constraint,
which prevents the parameter matrix from being very ill-conditioned.

%%这是因为可分离转换结构在降低模型参数量的同时也对参数矩阵条件数进行了限制，这不会使参数矩阵变得病态，还会提高模型对噪声和对抗攻击的防御能力
%This is because Separable Structured Transformations also limits the condition number of the parameter matrix while reducing the model parameters, 
%this will not make the parameter matrix ill-conditioned and improve the model's ability to resist noise or adversarial attack. 
%%与此相反的是，如果使用普通剪枝方法，压缩比越高，模型参数矩阵越稀疏，秩也越低，这将会破坏模型的防御能力
%On the contrary, if the ordinary pruning method is used, the higher the compression ratio, the stronger the sparsity of the parameter matrix, the lower rank of the parameter matrix, which will make the parameter matrix ill-conditioned and destroy the defensive ability of the model.

In previous experiments, we tested \emph{ARLST} and other baselines against the PGD attack at certain fixed perturbation levels. 
Besides, we also tested all models against the FGSM attack, the Square Attack (SA) and the AutoAttack (AA) on CIFAR-$100$ dataset when the compression rate is $100\times$.
Table.~\ref{table_cifar_100} shows the advantages achieved by \emph{ARLST}.
% As shown in Table~\ref{table_cifar_100}, 
% when the compression ratio is $100\times$, 
% the VGG-$16$-\emph{ARLST} has a small advantages.
%
% As we can see from Table~\ref{table_ImageNet}, Table~\ref{table_cifar_100}, Table~\ref{table_adv_fgsm}.
% The ImageNet is a large-scale dataset, and the adversarial training is time-consuming. 
%Therefore, it is unaffordable that conduct a complete experiment.
% we did implement the complex experiment on ImageNet. 
%As shown in Table~\ref{table_ImageNet}
Table.~\ref{table_ImageNet} gives an simple comparison to show the better performance of \emph{ARLST} on ImageNet.
\begin{table}[!ht]\small
    \centering
    \setlength\tabcolsep{5pt}
    \vspace{-2mm}
	\caption{\label{VGG_RLST_results}
	% Averaged classification comparison 
	Comparison between a VGG-16 and its variants on CIFAR-$100$ under adversarial training with $4$ attacks. 
% 	under adversarial training with different attack and the compression ratio being $100 \times$.
% 	The NA and RA of each algorithm under different attack on CIFAR-$100$ and the compression ration is $100 \times$. 
          }
    \begin{tabular}{ccccc}
    \hline
                 & HYDRA & ADMM & ATMC & ARLST \\ \hline
   Attacks            & \textbf{NA}/\textbf{RA}   & \textbf{NA}/\textbf{RA}   & \textbf{NA}/\textbf{RA}      & \textbf{NA}/\textbf{RA}       \\ \hline
    FGSM          & $44.9/29.7$ &$41.1/27.7$&$46.2/31.5$& $\textbf{47.1/32.1}$\\ 
    PGD           &$33.3/19.9$&$32.2/17.7$ &$36.6/21.3$& $\textbf{37.6/22.1}$ \\ 
    SA &  $43.0/31.7$ &$41.9/30.3$ &$42.9/33.6$& $\textbf{44.1/34.7}$    \\ 
    AA    & $31.9/17.7$& $30.9/16.9$ &$35.0/19.9$& $\textbf{35.2/20.8}$  \\ \hline
    \end{tabular}
    \label{table_cifar_100}
    \vspace{-4mm}
\end{table}
%%%%%%%%%%%%%%%%%%%%%%%%%%%%%%%%%%%%%%%%%%%%%%%%%%%%%%%%%%%%%%%%%%%%%%%%%%%%%%%%%%%%%%%%
%%%%%%%%%%%%%%%%%%%%%%%%%%%%%%%%%%%%%%%%%%%%%%%%%%%%%%%%%%%%%%%%%%%%%%%%%%%%%%%%%%%%%%%%
\begin{table}[!ht]\small
    \centering
	\caption{\label{VGG_RLST_results_imagenet}
		Comparison between a VGG-16 and its variants on ImageNet-1K under adversarial training with FGSM attack.
% 		The NA and RA of each algorithm under FGSM attack on ImageNet. 
		}
    \begin{tabular}{ccccc}
    \hline
                 & HYDRA & ADMM & ATMC & ARLST \\ \hline
       \textbf{Cr} & \textbf{NA}/\textbf{RA}       & \textbf{NA}/\textbf{RA} & \textbf{NA}/\textbf{RA}      & \textbf{NA}/\textbf{RA}       \\ \hline
    $5\times$         & $43.2/28.7$ &$41.2/27.9$&$45.1/29.5$&$\textbf{45.4/30.0}$   \\ 
    $10\times$           &$41.6/27.0$&$40.1/26.0$&$42.9/27.8$ &$\textbf{43.9/28.7}$    \\ \hline

    \end{tabular}
    \label{table_ImageNet}
\end{table}
\subsection{Vision Transformer}
To our best knowledge, there have been few existing studies on examining the robustness and Lightweight of ViT \cite{dosovitskiy2020image_iclr}.
Because ViT model training and adversarial training are very time-consuming,
we conduct a simple set of experiments based on ViT-B/16, comparing our method with HYDRA, ADMM, and ATMC on small datasets, Yale$-B$ and MNIST.
For adversarial training of ViT, we apply the FGSM attack to generate adversarial samples.
We set the perturbation magnitude to be 0.3 for MNIST and 0.03 for Yale$-$B.
We train four models for 100, 300 epochs on MNIST and Yale$-$B, respectively.

As shown in Table.~\ref{table_vit}, when the compression ratio is $3\times$, \emph{ARLST} achieves better performance.
This is because we have reduced the parameters of the FC layer and fully retained the Self-Attention, such that the performance of the model is not severely damaged.

%%%%%%%%%%%%%%%%%%%%%%%%%%%%%%%%%%%%%%%%%%%%%%%%%%%%%%%%%%%%%%%%

%-------------------------------------------------------------------------

\section{Conclusions}
\label{sec:05}
%
%In this work, we have 
This work proposes a learning model with the properties of both \textit{lightweight} and \textit{robustness}. We approximate the weight matrix of each linear layer by the tensor product of several separable structured small-sized matrices. Furthermore, we propose a joint constraint of sparsity and differentiable condition number which is imposed on these separable matrices, with the goal of promoting both the \textit{lightweight} and the \textit{robustness} of the whole system.
%
%Since the Kronecker product linear system has the advantages of sparsity maintenance, as well as
%rank and condition number maintenance, 
% Based on the advantages of sparsity and condition number,
% we imposed the regularization term, which aims to improve the sparsity and obtain the appropriate condition number, on each separable matrix to promote both the \textit{lightweight} and the \textit{robustness} of the whole system.
%
We conduct min-max adversarial training based on several models combined with the proposed method, namely \textit{ARLST}, to defend against adversarial attacks.
The experimental results of MLP, VGG-16 and ViT show that \emph{ARLST} outperforms the SOTA method based on the original fully-connected layers, achieving higher robustness to various adversarial perturbations while having fewer network parameters and less risk of overfitting.
% In future, 
Moreover, the proposed \emph{ARLST} model is flexible and can be extended to other cases of 
deep learning models, such as the attention module and the graph neural networks.

%%%%%%%%% REFERENCES
{\small
\bibliographystyle{ieee_fullname}
\bibliography{egbib}
}

\clearpage

\appendix

%%%%%%%%% BODY TEXT - ENTER YOUR RESPONSE BELOW
\section{Notations and Definitions}
In the submitted manuscript and supplementary material, vectors, matrices, tensors are denoted by bold lower case letters $\mathbf{v}$, bold upper case ones $\bm{V}$, 
and mathcal letter ${\mathcal{V}}$, respectively.
%Matrices are written as boldface capital letters like $\X$, $\Phib$, 
%column vectors are denoted by boldfaced small letters, e.g. $\x$, 
%$\d$, whereas scalars are either capital or small letters, such as $n$ and $N$. 
% $\bm{I}_n$ the $n\times n-$identity matrix, 
%$\mathbf{1}_k$ the $k$ dimensional all-ones column vector,
%$\mathbf{1}_{k\times k}$ the $k\times k$ all-ones square matrix, 
% $\|\cdot\|_F$ denotes  the Frobenius norm of matrices,  
%the operations ${v} \cdot {w}$, $\mathbf{v} \cdot \mathbf{w}$, $\mathbf{V} \cdot \mathbf{W}$
%denote the product of two scalars, vectors, and matrices, respectively,
$\mathbf{v} \in \mathbb{R}^{ab}$ denotes a vector $\mathbf{v}$ with the dimension $ab$.
$\mathbf{V} \in \mathbb{R}^{a\times b}$ denotes a matrix $\mathbf{V}$ with the resolution $a\times b$.
${V}_i$, ${V}_{ij}$, ${V}_{i_1 i_2 \cdots i_T}$ denotes the $i^{th}$ column of matrix $\mathbf{V}$,
the entry in the $i^{th}$ row and the  $j^{th}$ column of $\mathbf{V}$, 
the entry in a tensor ${\mathcal{V}}$ with indices $i_j$ indicating the position in the respective mode.
%
%%%%%%%%%%%%%%%%%%%%%%%%%%%%%%%%%%%%%%%%%%%%%%%%%%%%%%%%%%%%%%%%%%%%%%%%%%
\section{Proof of Eq.~(9) in the Submitted Manuscript}
In the submitted manuscript, the Eq.~(9) is shown as the following  Eq.~\eqref{eq_condition_number}
\begin{equation} 
	\label{eq_condition_number}
	\kappa(\W) = \kappa(\mathbf{A}^{(1)} ) \kappa(\mathbf{A}^{(2)} ) \cdots \kappa(\mathbf{A}^{(T)} ).
\end{equation}

Before the proof, we refer to a large, structured transformation matrices $\W$
that can be represented as a concatenation of smaller matrices
$\mathcal{A} := \{\mathbf{A}^{(1)}, \mathbf{A}^{(2)}, \cdots, \mathbf{A}^{(T)}\}$
as a separable transformation,
\begin{equation}
	\label{eq_matrix_kronecker_decomposition}
	\W = \mathbf{A}^{(1)} \otimes \mathbf{A}^{(2)} \otimes \cdots \otimes \mathbf{A}^{(T-1)} \otimes \mathbf{A}^{(T)},
\end{equation}
with the properties \cite{graham2018kronecker_book} of 
\begin{align}
	\begin{split}
		\label{eq_matrix_kronecker_decomposition_exchange}
		\mathbf{A}^{(1)} \otimes \mathbf{A}^{(2)} \otimes \mathbf{A}^{(3)} 
		& = (\mathbf{A}^{(1)} \otimes \mathbf{A}^{(2)}) \otimes \mathbf{A}^{(3)} \\
		& = \mathbf{A}^{(1)} \otimes ( \mathbf{A}^{(2)} \otimes \mathbf{A}^{(3)} ),
	\end{split}
\end{align}
\begin{equation}
	\label{eq_matrix_kronecker_norm}
	\|\mathbf{A}^{(1)} \otimes \mathbf{A}^{(2)} \otimes \mathbf{A}^{(3)}\|
	= \|\mathbf{A}^{(1)}\| \|\mathbf{A}^{(2)}\| \|\mathbf{A}^{(3)}\|,
\end{equation}
and 
\begin{equation}
	\label{eq_matrix_kronecker_decomposition_rank}
	\text{rank}(\W) = \text{rank}(\mathbf{A}^{(1)})\text{rank}(\mathbf{A}^{(2)})\cdots\text{rank}(\mathbf{A}^{(T)}).
\end{equation}

\begin{theorem}[Theorem $4.2.15$ in \cite{horn1994topics_book}]
	\label{theorem:01}
	Let $\A \in \mathbb{R}^{a\times b}$, $\B \in \mathbb{R}^{c\times d}$ have rank
	$\text{rank}(\A)$, $\text{rank}(\B)$, and let 
	$\mathbf{V}_{\A}, \mathbf{W}_{\A}, \bf{\Sigma}_{\A}, \mathbf{V}_{\B}, \mathbf{W}_{\B}, \bf{\Sigma}_{\B},$ 
	be the matrices corresponding to their
	singular value decompositions. Then we can easily derive the singular value
	decomposition of their Kronecker product:
	\begin{equation}
		\label{eq_svd}
		\begin{split}
			\A \otimes \B &  = ( \mathbf{V}_{\A} \bf{\Sigma}_{\A} \mathbf{W}_{\A}^\top ) \otimes 
			( \mathbf{V}_{\B} \bf{\Sigma}_{\B} \mathbf{W}_{\B}^\top ) \\
			& = ( \mathbf{V}_{\A} \otimes  \mathbf{V}_{\B}) ( \bf{\Sigma}_{\A} \otimes \bf{\Sigma}_{\B})
			( \mathbf{W}_{\A} \otimes  \mathbf{W}_{\B})^\top.
		\end{split}
	\end{equation}
\end{theorem}

In Eq.~\eqref{eq_svd}, the singular values of 
$\mathbf{V}_{\A}, \mathbf{W}_{\A}, \mathbf{V}_{\B}, \mathbf{W}_{\B}$ are equal to $1$.
$\bf{\Sigma}_{\A}$ and $\bf{\Sigma}_{\B}$ are diagonal and contain the singular values of $	\A$ and $\B$.
The condition number of $\A \otimes \B$ could be rewritten as
\begin{align*}
	\label{eq_condition_number_proof}
	k(\A \otimes \B) &  =k( ( \mathbf{V}_{\A} \otimes  \mathbf{V}_{\B}) ( \bf{\Sigma}_{\A} \otimes \bf{\Sigma}_{\B})
	( \mathbf{W}_{\A} \otimes  \mathbf{W}_{\B}) ) \\
	&  = k( \mathbf{V}_{\A} \otimes  \mathbf{V}_{\B}) k( \bf{\Sigma}_{\A} \otimes \bf{\Sigma}_{\B}) 
	k( \mathbf{W}_{\A} \otimes  \mathbf{W}_{\B}) \\
	&  = k( \bf{\Sigma}_{\A} \otimes \bf{\Sigma}_{\B}) \\
	&  = \dfrac{\sigma_{\max} (\bf{\Sigma}_{\A}) \sigma_{\max} (\bf{\Sigma}_{\B})}{\sigma_{\min} (\bf{\Sigma}_{\A})\sigma_{\min} (\bf{\Sigma}_{\B})} \\ 
	&  = k(\A)k(\B)
\end{align*}

Recalling the property in Eq.~\eqref{eq_matrix_kronecker_decomposition_exchange}, it is easy to 
conclude the Eq.~\eqref{eq_condition_number}
if Eq.~\eqref{eq_matrix_kronecker_decomposition} holds.

%
%%%%%%%%%%%%%%%%%%%%%%%%%%%%%%%%%%%%%%%%%%%%%%%%%%%%%%%%%%%%%%
%
\section{A 2D Example}
In the following, the problem \eqref{eq_basic_linear_system} could be easily rewritten into the large-scale form of Eq.~\eqref{eq_separable_multi_layer_tensor}.
For example, by regarding $\x \in \mathbb{R}^m$, $\y \in \mathbb{R}^{d}$, 
$\W \in \mathbb{R}^{d\times m}$ in Eq.~\eqref{eq_basic_linear_system}, 
its two-dimensional system with separable structured transformations can be rewritten as
%%%%%%%%%%%%%%%%%%%%%%%%%%%%%%%%%%%%%%%%%%%%%
\begin{equation}
	\label{eq_separable_layer}
	\mathbf{Y} = \mathbf{A}\mathbf{X}\mathbf{B}^{\top},
	% Y = \sigma(AXB^T + b)
\end{equation}
%%%%%%%%%%%%%%%%%%%%%%%%%%%%%%%%%%%%%%%%%%%%
%%%%%%%%%%%%%%%%%%%%%%%%%%%%%%%%%%%%%%%%%%%%
\begin{equation}
	\label{eq_basic_linear_system}
	\y = \W \x, \;\;\;  \mathbf{h}  = \sigma(\y)
	% \y = \W\circ \x %\y = \W\x +\bm{b}
\end{equation}
%%%%%%%%%%%%%%%%%%%%%%%%%%%%%%%%%%%%%%%%%%%%
\begin{equation}
	\begin{aligned}
		\label{eq_separable_multi_layer_tensor}
		%\text{vec}(\mathcal{Y}) = \sigma\left( \left( \mathbf{A}^{(T)} \otimes \cdots \otimes \mathbf{A}^{(2)} \otimes \mathbf{A}^{(1)}\right) \text{vec}(\mathcal{X}) + \mathbf{c} \right) 
		\text{vec}(\mathcal{Y}) =  \left( \mathbf{A}^{(1)} \otimes \mathbf{A}^{(2)} \otimes \cdots \otimes \mathbf{A}^{(T-1)} \otimes \mathbf{A}^{(T)}\right) \text{vec}(\mathcal{X}),
		%& = \sigma\left(  \text{vec} \left( {\A}^{(1)} {\text{vec}}^{-1}(\x) {\left(\mathbf{A}^{(T)} \otimes \cdots \otimes \mathbf{A}^{(2)}\right)}^\top  \right) + \mathbf{c}  \right),
		%& = \sigma\left(  \text{vec} \left( {\A}^{(1)} \X {\left(\mathbf{A}^{(T)} \otimes \cdots \otimes \mathbf{A}^{(2)}\right)}^\top  \right) + \mathbf{c}  \right)
	\end{aligned}
	% y = \sigma(B $\otimes$ A)x + b
\end{equation}
%%%%%%%%%%%%%%%%%%%%%%%%%%%%%%%%%%%%%%%%%%%%
%%%%%%%%%%%%%%%%%%%%%%%%%%%%%%%%%%%%%%%%%%%%%%%%%%%%%%%
\begin{equation}
	\label{eq_basic_linear_trans_MD}
	%\y = \sigma \left(\W\x + \mathbf{c} \right)
	%\mathbf{Y} = \sigma\left( \mathbf{A}\mathbf{X}\mathbf{B}^{\top} + \mathbf{A} \mathbf{C} \mathbf{B}^{\top} \right) 
	%\mathcal{Y} = \sigma\left( \mathcal{X} \times_{1}\mathbf{A}^{(1)}  \times_{2}\mathbf{A}^{(2)} \times_{3} \cdots  \times_{T} \mathbf{A}^{(T)} +  \mathcal{C} \right). 
	\mathcal{Y} =  \mathcal{X} \times_{1}\mathbf{A}^{(1)}  \times_{2}\mathbf{A}^{(2)} \times_{3} \cdots  \times_{T} \mathbf{A}^{(T)}. 
\end{equation}
%%%%%%%%%%%%%%%%%%%%%%%%%%%%%%%%%%%%%%%%%%%%%%%%%%%%%
where $ \mathbf{A}\in \mathbb{R}^{K_1\times I_1}$ and $\mathbf{B}\in \mathbb{R}^{K_2\times I_2}$
have smaller $I_1$ and $I_2$.
$\mathbf{X} \in \mathbb{R}^{I_1\times I_2}$ and $\mathbf{Y}\in \mathbb{R}^{K_1\times K_2}$ are reshaped
two-dimensional matrices from $\x$ and $\y$. 
If we read $\W = \mathbf{A} \otimes \mathbf{B}$ with  $K_1 K_2 = d$ and $I_1 I_2 = m$,
% As shown from Eq.~\eqref{eq_basic_linear_system}, Eq.~\eqref{eq_separable_layer} and Eq.~\eqref{eq_Kronecker}, 
the dimension of the problem is reduced from $K_1K_2I_1I_2$ in Eq.~\eqref{eq_basic_linear_system}
to $K_1I_1 + K_2I_2$ in Eq.~\eqref{eq_separable_layer}.
%  from Eq.~\eqref{eq_basic_linear_system} to Eq.~\eqref{eq_separable_layer}. 
The ratio of the parameters of two models is computed by $\eta_1 = \frac{1}{K_1I_1} + \frac{1}{K_2I_2}$.
%%
%\begin{equation}
%	\label{eq_ratio_model_size}
%	\eta_1 = \frac{1}{K_1I_1} + \frac{1}{K_2I_2}.
%	% Y = \sigma(AXB^T + b)
%\end{equation}
%%
The latter is much smaller than the former following the increase of $K_1I_1$ and $K_2I_2$.

Considering the matrix multiplication in Eq.~\eqref{eq_separable_layer}, 
the computational complexity is also significantly reduced from 
$\mathcal{O}(\W\x) = \mathcal{O}(K_1\times I_1 \times I_2 \times K_2)$ to
$\mathcal{O}(\mathbf{A}\mathbf{X}\mathbf{B}^{\top}) = \mathcal{O}(K_1\times I_1 \times I_2 + K_1\times I_2 \times K_2)$,
and the ratio of them is computed as $\eta_2 =  \frac{1}{K_2} + \frac{1}{I_1}$.
%%
%\begin{equation}
%	\label{eq_ratio_compute_complexity}
%	\eta_2 =  \frac{1}{I_1} + \frac{1}{I_2}.
%\end{equation}
%%
%or Eq.~\eqref{eq_separable_multi_layer}.
% The compression ratio $\eta$ of them is expressed by the division of uncompressed model size by compressed size.
The reduction of computational complexity depends on the sizes of $\X$ and $\Y$, i.e., $I_1$ of $\X$ and $K_2$ of $\Y$ with $d, m$ being fixed.

Eq.~\eqref{eq_basic_linear_trans_MD} shows that the number of parameters $\mathcal{A}$ and its computational complexity
%It is easy to conclude that the number of parameters $\mathcal{A}$ and the computational complexity in Eq.~\eqref{eq_basic_linear_trans_MD}
depend on the size of input tensor $\mathcal{X}$ and the size of output tensor $\mathcal{Y}$.
The constructions of $\mathcal{X}$ and $\mathcal{Y}$ determine the structure of $\mathcal{A}$ and the computational load.
\section{ARLST with Adversarial Training}
In the submitted manuscrip, VGG-$16$-\emph{ARLST} outperforms ATMC, ADMM and HYDRA
in CIFAR-10 and SVHN dataset under different levels of compression ratio.
As shown in Table~\ref{table_adv_fgsm},
by using the same pre-trained setting, VGG-$16$-\emph{ARLST} outperforms ATMC, ADMM and HYDRA
in CIFAR-10 datasets under adversarial training with FGSM attack.
It is worth noting that the VGG-$16$-\emph{ARLST} achieves overwhelming advantage when the compression ratio is very high, e.g., $100$ times. % By taking ResNet-$5$ performed on MNIST as an example, Fig.~\ref{fig_condition_number2} 
Additionally, the VGG-$16$-\emph{ARLST} has a smaller variance of the accuracies for five runs in the experiments.
These results suggest that the VGG-$16$-\emph{ARLST} has higher stability than above SOTA methods. 

\begin{table*}[h!]\small
    \centering
    \caption{\label{table_adv_fgsm} 
 		Comparison between a VGG-16 and its variants under adversarial training with FGSM attack.
 	    }	
    \begin{tabular}{cccccc}
    \hline
     & & VGG16-HYDRA & VGG16-ADMM &VGG16-ATMC & VGG16-ARLST       \\ \hline
     & \textbf{Cr} & \textbf{NA}/\textbf{RA} (\textbf{Var})  &  \textbf{NA}/\textbf{RA}(\textbf{Var}) &  \textbf{NA}/\textbf{RA}(\textbf{Var}) & \textbf{NA}/\textbf{RA}(\textbf{Var})  \\ \hline
    \multirow{6}{*}{\begin{tabular}{@{}c@{}} \rotatebox[origin=c]{90}{CIFAR-10} \end{tabular}} 
    &$1\times $(pretrain, \textbf{NAT}) & $96.2/22.4$ & $96.2/22.4$ & $96.2/22.4$ & $96.2/22.4$ \\ 
    & $1 \times $(pretrain, \textbf{AT})  & $84.1/69.4(0.23/0.15)$ & $82.3/64.5(0.40/0.25)$ &$83.2/64.1(0.22/0.09)$& $\textbf{85.2/70.4}(0.28/0.12)$ \\ 
    & $10 \times $  & $81.3/67.1(0.18/0.50)$ & $81.1/61.6(0.22/0.32)$ &$80.5/60.8(0.29/0.20)$& $\textbf{83.8/69.8}(0.23/0.20)$ \\ 
    & $20 \times $    & $81.2/66.5(0.25/0.33)$ & $78.5/59.9(0.20/0.35)$ &$78.9/59.7(0.19/0.25)$& $\textbf{82.6/68.4}(0.10/0.12)$ \\ 
    & $100 \times $    & $69.9/53.5(0.25/0.35)$ & $65.3/48.9(0.23/0.34)$ &$66.1/50.0(0.32/0.15)$& $\textbf{74.5/58.3}(0.12/0.09)$ \\ 
    & $200 \times $    & $45.1/37.4(0.19/0.20)$ & $20.5/10.3(0.35/0.46)$ &$61.9/43.2(0.29/0.25)$& $\textbf{63.9/49.1)}(0.20/0.12$ \\ \hline
    \end{tabular}
\end{table*}

% \section{More Network Architure Detail}
\section{The Application in CNNs}
% \begin{figure}[!ht]
% 	\centering
% 	\includegraphics[width=0.5\textwidth]{picture/asymmetric convolution.png}
% 	\caption{
% 		\label{fig_cnn}
% 		%
% 		Traditional convolution and separable asyminetric convolution
% 	}
% \end{figure}
In this section, we construct the \emph{RLST} function and the \emph{ARLST} function for learning robust and lightweight models
% GNNs, and Attention 
for classic CNNs. 
The following two parts in a common CNN can be modified by the proposed operator defined in Eq.~(2) of the Submitted Manuscript.

%%%%%%%%%%%%%%%%%%%%%%%%%%%%%%%%%%%%%%%%%
%\romannumeral 2) The FC layers associated with the Flatten layer. 
\paragraph{The FC layers}
In most CNNs, the input before flatten is often a $3$D $(\text{height, width, number of filters})$ tensor,
and the Flatten operator reshapes the $3$D tensor to have a shape that is equal to the number of elements contained in the tensor.
% unrolls the $3$D tensor to a high-dimensional vector. 
Differently, the proposed tensor operator can directly operate the $3$D input tensor and remove the Flatten layer.
Hence, all classical FC layers become the tensor multiplication as shown in Eq.~(2) of the Submitted Manuscript.
It is worth noticing that the proposed operation is adapted to the structures of input tensor without flattening,
which is different to the work in \cite{wu2016compression_icaci}.
The latter used Kronecker product decomposition to approximate huge FC weight matrices after flatten layer.
The similar methods based on Kronecker tensor decomposition are also shown in 
\cite{cohen2016expressive_clt, budden2017deep_icml, chen2019compressing_iccsnt, novikov2015tensorizing_nips}.
%This is also different the traditional tensor decomposition 
The computation for tensor decomposition is expensive when processing high-dimensional signals.

By taking ZF$-5$ as an example performed on Extended Yale$-$B dataset, 
we follow its basic architecture and construct \emph{ARLST-}ZF$-5$ network.
The input before flatten is a $3$D tensor with the size of $7\times 7\times 64$.
We then remove the flatten layer of ZF$-5$ and reshape the $3$D tensor as $49\times 64$.
Thus, each following FC layer could be constructed by two separable transformations, i.e., $T = 2$ in 
Eq.~(2) of the Submitted Manuscript, and details are shown in Eq.~(6) of the Submitted Manuscript.
For example, in the FC1 layer, \emph{ARLST-}ZF$-5$ uses $\A^{(1)} \in \mathbb{R}^{64\times 49}$ and $\A^{(2)} \in \mathbb{R}^{64\times 64}$,
to replace a $\mathbb{R}^{3136 \times 1024}$ transformation in original ZF$-5$ network.
Hence, in the FC2 layer, \emph{ARLST-}ZF$-5$ uses two $\mathbb{R}^{64\times 64}$ matrices to 
replace a $\mathbb{R}^{1024 \times 1024}$ transformation in original ZF$-5$ network.
Finally, \emph{ARLST-}ZF$-5$ uses $\A^{(1)} \in \mathbb{R}^{1\times 64}$ and 
$\A^{(2)} \in \mathbb{R}^{38\times 64}$ to project the tensor to a $38-$dimensional vector,
by replacing a $\mathbb{R}^{1024 \times 38}$ transformation in original ZF$-5$ network.
%An intuitive example is shown in Fig.~\ref{fig_tensor_example}.
%
% \textcolor{red}{
The three FC layers of ZF$-5$ have $14,534,294$ parameters, but \emph{ARLST-}ZF$-5$ 
only has $259,606$ parameters, and $98.21\%$ parameters have been reduced. 
%}
%The three FC layers of ZF$-5$ have $4,298,752$ parameters, 
%but \emph{ARLST-}ZF$-5$ only has $17,920$ parameters, and $99.58\%$ parameters have been reduced.
%
%\begin{figure}[!ht]
%	\centering
%	\includegraphics[width=0.5\textwidth]{fig/xiaodi/Asymmetric_convolution1.png}
%	\caption{
%		\label{fig_cnn_kc}
%		%
%		Traditional convolution and deep separable asyminetric convolution
%	}
%\end{figure}
%%

%%%%%%%%%%%%%%%%%%%%%%%%%%%%%%%%%%%%%%%%%%%%%%%%%%%%
\begin{figure}[!ht]
	\centering
	\includegraphics[width=0.45\textwidth]{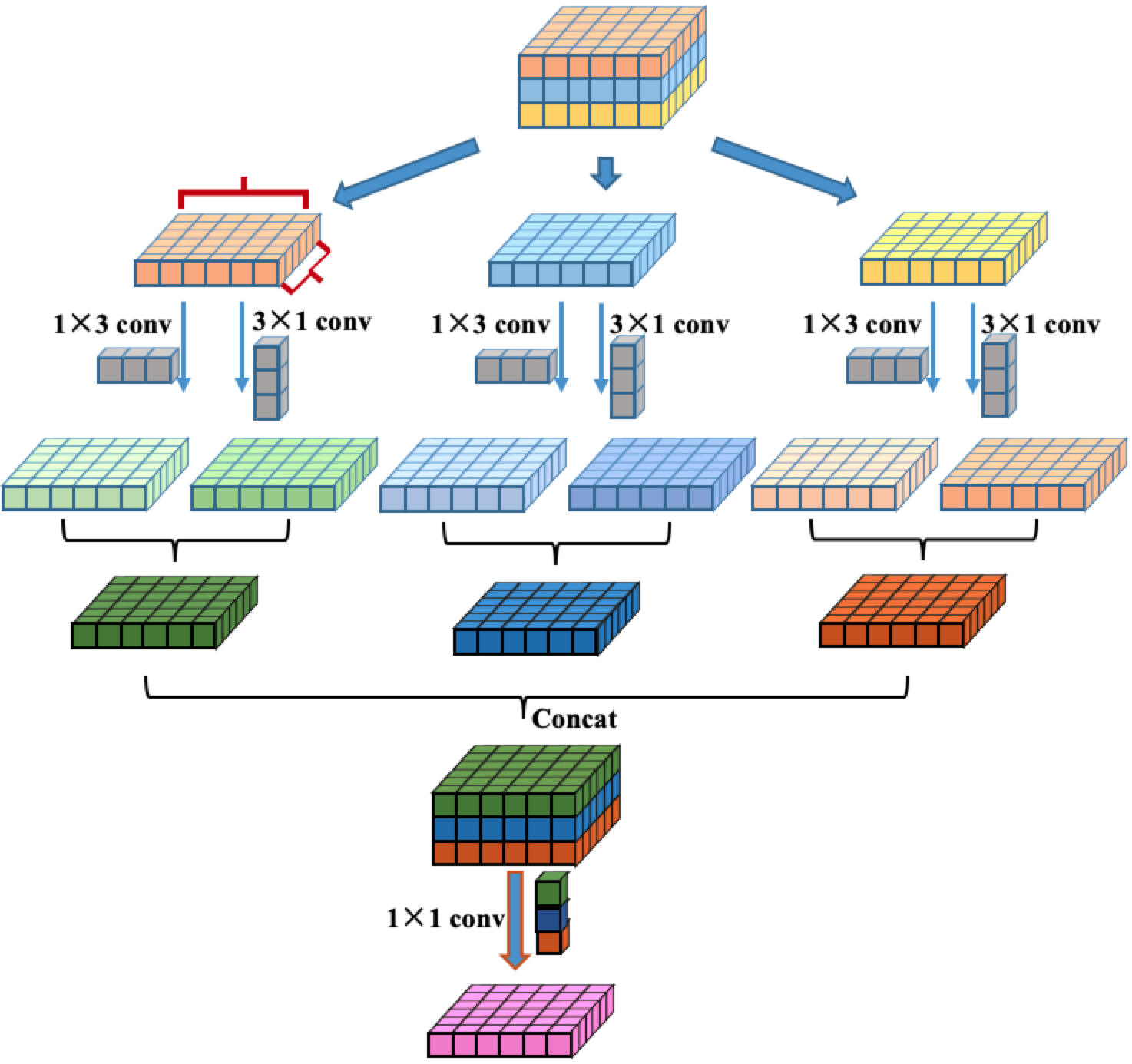}
	\caption{
		\label{fig_cnn_kc}
		This figure depicts a way for constructing the convolution layer by using separable transformations.
	}
\end{figure}
%%%%%%%%%%%%%%%%%%%%%%%%%%%%%%%%%%%%%%%%%%%%%%%%%%%%%%%%%%%
% \begin{table}[htb!]
% 	\caption{Network structure of feature extraction module}
% 	\centering
% 	\begin{tabular}{p{1.3cm}p{1.33cm}p{1.3cm}p{1.3cm}p{1.0cm}}
% 		\hline
% 		ID &    Filter  & Channel & Strides & Act  \\
% 		\hline
% 		Conv1  & [3,1][1,3]  &16   & (1,1)  & ReLu      \\
% 		Conv2  & [3,1][1,3]  & 16  & (1,1)  & ReLu      \\
% 		Pool3  & [2,2]  & ----  & (2,2)  & ----        \\
% 		Conv4  & [3,1][1,3]  & 32  & (1,1)  & ReLu      \\
% 		Conv5  & [3,1][1,3]  & 32  & (1,1)  & ReLu      \\
% 		Pool6  & [2,2]   & ----  & (2,2)  & ----        \\
% 		Conv7  & [3,1][1,3]   & 64  & (1,1)  & ReLu      \\
% 		Conv8  & [3,1][1,3]   & 64  & (1,1)  & ReLu      \\
% 		Conv9  & [3,1][1,3]  & 64  & (1,1)  & ReLu      \\
% 		Pool10 & [2,2]  & ----  & (2,2)  & ----        \\
% 		\hline       
% 	\end{tabular}
% 	\label{table_structure}
% \end{table}
\begin{table}[htb!]
	\caption{\label{table_net_structure_vgg9_convlayer}
		Network structure of convolution layers.}
	% \vspace{20pt}
	\centering
	\begin{tabular}{p{1.3cm}p{1.33cm}p{1.3cm}p{1.3cm}p{1.3cm}p{1.2cm}}
		\hline
		ID & Filter size & Num & Strides & $Act  $ \\
		\hline
		Conv1-1  & [3,1][1,3]  &16   & (1,1)  & ----      \\
		Conv1-2  & [1,1]  &  16   & (1,1)  & ReLu      \\
		Conv2-1  & [3,1][1,3]   & 16  & (1,1)  & ----       \\
		Conv2-2  & [1,1]   & 16  & (1,1)  & ReLu      \\
		Pool3  & [2,2]  & --  & (2,2)  & --        \\
		Conv4-1  & [3,1][1,3]  & 32  & (1,1)  & ----       \\
		Conv4-2  & [1,1]  & 32  & (1,1)  & ReLu      \\
		Conv5-1  & [3,1][1,3]  & 32  & (1,1)  & ----      \\
		Conv5-2  & [1,1]  & 32  & (1,1)  & ReLu      \\
		Pool6  & [2,2]   & --  & (2,2)  & --        \\
		Conv7-1  & [3,1][1,3]   & 64  & (1,1)  & ----      \\
		Conv7-2  & [1,1]  & 64  & (1,1)  & ReLu      \\
		Conv8-1  & [3,1][1,3]   & 64  & (1,1)  & ----       \\
		Conv8-2  & [1,1]   & 64  & (1,1)  & ReLu      \\
		Conv9-1  & [3,1][1,3]  & 64  & (1,1)  & ----      \\
		Conv9-2  & [1,1] & 64  & (1,1)  & ReLu      \\
		Pool10 & [2,2]  & --  & (2,2)  & ----        \\
		\hline       
	\end{tabular}
	\label{table_structure}
\end{table}

% In order to further reduce the amount of parameters, we also design a separable feature extraction module on the convolution layer.the figure above\ref{fig_cnn} ,is used  a traditional way of 3 × 3 convolution complete feature extraction process, the figure below \ref{fig_cnn}, and directly with two asymmetric convolution kernels 1×3 and 3×1 to replace the original 3 × 3 convolution.
% It is easy to conclude that this feature extraction method can reduce the calculation amount by 33\%. By balance the number of channels, network depth, and combining with LRST module, we finally get VGG10-LRST and VGG16-LRST. The VGG10 feature extraction module is shown in the table\ref{table_structure}.

\paragraph{The Asymmetric Convolutional Layers}
Since the fully-connected layer can be treated as a kind of $1 \times 1$ convolution, 
we expand the work to the convolutional layers. 
%\textcolor{red}{ For the traditional convolution filter, the size of the convolution kernel can be expressed as 
%	$\mathcal{X}_i \in\mathbb{R}^{k_1\times k_2\times c}$.}
Given an input of a convolutional layer with $c$ channels $\mathcal{I} \in \mathbb{R}^{p\times q \times c}$, 
all patches set can be written as $\mathcal{X} = [\mathcal{X}_1, \mathcal{X}_2, \cdots, \mathcal{X}_n]$
with the $i^{\text{th}}$ patch  $\mathcal{X}_i \in\mathbb{R}^{k_1\times k_2\times c}$.
Considering a $c$-dimensional traditional convolutional layer
with a filter size of $k_1\times k_2 \times c$ to convolute the image patch $\mathcal{X}_i$, 
where $k_1, k_2$ are the spatial size of the filter. Let $n$ denote the number of output channels.
%Let $C_{in}^{j}$ denote the number of filters in $j^{\text{th}}$.
% All patches set can be written as $\mathcal{X} = [\mathcal{X}_1, \mathcal{X}_2, \cdots, \mathcal{X}_n]$.
%
% We use $\x_i \in \mathbb{R}^{c k_1 k_2}$ to denote a vector that reshapes this volume
% and $\X = [\x_1, \x_2, \cdots, \x_n] \in \mathbb{R}^{(c k_1 k_2) \times n}$ to denote a matrix containing all patch vectors.
%
Traditionally, let $\A\in \mathbb{R}^{k_1\times k_2 \times c}$ be a square kernel, and $\A \ast \mathcal{X}_i$ is a convolution operation.
In the work, we construct the asymmetric convolutional layer,
as shown in Fig.~\ref{fig_cnn_kc}.
It replaces the square-kernel convolution operation.
by separable transformations in three directions, i.e., $\A^{(1)}\in \mathbb{R}^{ k_1 \times 1 \times 1}$ with $c$ channels, 
$\A^{(2)}\in  \mathbb{R}^{1\times k_2 \times 1}$ with $c$ channels,
and $\A^{(3)}\in  \mathbb{R}^{1\times 1 \times c}$ with $n$ channels, 
as depicted in Fig.~\ref{fig_cnn_kc}.
The Kronecker product of $\A^{(1)}, \A^{(2)}, \A^{(3)}$ forms a tensor $\A \in \mathbb{R}^{ k_1 \times k_2 \times c}$.

This is different from the depthwise separable convolution,
which divides the standard filter with the size of $k_1\times k_2 \times c \times n$ into two separable filters with the sizes of
$1 \times k_1 \times k_2 \times c$ and $1\times 1 \times c \times n$, respectively. 
Differently, The proposed method has the smaller parameter sizes and lower computational cost.
For one layer of the proposed convolution and the depthwise separable convolution, the ratio of parameters, 
and the ratio of computational complexity are given by
\begin{align}
% \label{eq_ratio_model_size}
\eta_1 & = \frac{k_1 c + k_2 c + cn }{k_1 k_2 c + c n} =  \frac{k_1  + k_2  + n }{k_1 k_2  +  n}, \\
\eta_2 & =  \frac{k_1 c pq + k_2 c pq + cnpq}{k_1 k_2 p q c + c n pq} =  \frac{k_1  + k_2  + n }{k_1 k_2  +  n}.
% Y = \sigma(AXB^T + b)
\end{align}
The numerator is smaller than the denominator following the increase of $k_1$ and $k_2$.
Additionally, compared with the methods based on asymmetric convolutions, 
e.g. ACNet \cite{ding2019acnet_iccv} and Inception V2 \cite{szegedy2016rethinking_cvpr},
%compared with ACNet \cite{ding2019acnet_iccv}, 
the proposed method cancels the square-kernel convolution.

% %%%%%%%%% REFERENCES
% {\small
% \bibliographystyle{ieee_fullname}
% \bibliography{egbib}
% }

\end{document}